\documentclass[preprint,12pt]{elsarticle}

\usepackage{amssymb}

\usepackage[table,xcdraw]{xcolor}

\usepackage[utf8]{inputenc} 
\usepackage[T1]{fontenc}    
\usepackage{hyperref}       
\usepackage{url}            
\usepackage{booktabs}       
\usepackage{amsfonts}       
\usepackage{nicefrac}       
\usepackage{amstext}
\usepackage{booktabs}
\usepackage{multirow}
\usepackage{algorithmic}
\usepackage{algorithm,amsmath}

\usepackage{tikz}
\usepackage{lscape}
\usepackage{todonotes}

\usepackage{graphicx}
\usepackage{subcaption}

\usepackage{paralist}

\usetikzlibrary{shapes,arrows,positioning}

\journal{Journal of Manufacturing Systems}
\definecolor{red}{RGB}{0, 0, 0}

\begin{document}

\begin{frontmatter}



\title{Predicting machine failures from multivariate time series: an industrial case study}


\author[inst1]{Nicolò Oreste Pinciroli Vago}
\ead{nicolooreste.pinciroli@polimi.it}

\author[inst1]{Francesca Forbicini}
\ead{francesca.forbicini@mail.polimi.it}

\affiliation[inst1]{organization={Department of Electronics, Information and Bioengineering, Politecnico di Milano},
            addressline={Via Giuseppe Ponzio, 34}, 
            city={Milan},
            postcode={20133}, 
            country={Italy}}

\author[inst1]{Piero Fraternali}
\ead{piero.fraternali@polimi.it}

\begin{abstract}
\textcolor{red}{Non-neural} Machine Learning (ML) and Deep Learning (DL) models are often used to predict system failures in the context of industrial maintenance. However, only a few researches jointly assess the effect of varying the amount of past data used to make a prediction and the extension in the future of the forecast. This study evaluates the impact of the size of the reading window and of the prediction window on the performances of models trained to forecast failures in \textcolor{red}{three data sets concerning the operation of (1)} an industrial wrapping machine working in discrete sessions\textcolor{red}{, (2) an industrial blood refrigerator working continuously, and (3) a nitrogen generator working continuously}. The problem is formulated as a binary classification task that assigns the positive label to the prediction window based on the probability of a failure to occur in such an interval. \textcolor{red}{Six} algorithms (logistic regression, random forest, support vector machine, LSTM, \textcolor{red}{ConvLSTM, and Transformers}) are compared using multivariate telemetry time series. The results indicate that, in the considered scenario\textcolor{red}{s}, \textcolor{red}{the dimension of  the prediction windows plays a crucial role} and highlight the effectiveness of DL approaches at classifying data with \textcolor{red}{diverse time-dependent patterns preceding a failure and the effectiveness of ML approaches at classifying similar and repetitive patterns  preceding a failure}.
\end{abstract}


\begin{highlights}
\item A comparison of \textcolor{red}{non-neural Machine Learning} and \textcolor{red}{Deep Learning} for failure prediction in \textcolor{red}{three industrial cases} is proposed
\item \textcolor{red}{We quantify the diversity of anomalous patterns preceding faults}
\item \textcolor{red}{Results reveal Deep Learning superiority over non-neural Machine Learning approaches for failure prediction only for complex data sets with more diverse anomalous patterns}
\item Increasing the amount of historical data does not necessarily yield better results
\end{highlights}


\end{frontmatter}


\section{Introduction}
Predictive maintenance is a strategy that aims to reduce equipment degradation and prevent failures \cite{Bousdekis2020, Zonta2020, Dalzochio2020} by forecasting their occurrence. It differs from corrective maintenance (in which the components are used for their entire life span and replaced only when there is a fault) \cite{SHEUT1994, Wang2014} and preventive maintenance (in which the replacement and reparation activities are scheduled periodically) \cite{Meller1996, ShaominWu2010}. 

Industrial data are often the result of the monitoring of several physical variables, which produces multivariate time series \cite{Liang2021, Tian2023}. Predicting failures via manual inspection is time-consuming and costly, especially because the correlation among multiple variables can be difficult to appraise with manual approaches. Computer-aided methods can improve performance and reliability and are less onerous and error-prone \cite{Salfner2010, Garca2010, Leukel2022}.

Predicting machine failures requires the  cycle of data collection, model identification, parameter estimation, and validation \cite{Box2016}. Alternative models are compared in terms of the effort needed to train them, the performances they deliver, and the relationship between such performances and the amount of input data used to make a prediction and the time horizon of the forecast.

In this  paper, we compare alternative models trained for predicting malfunctions.
The focus of the analysis is to study how performances, measured with the macro $F_1$ score metrics, are affected by the length of the input time series fragment used to make the prediction (\textit{reading window -- RW}) and by the extension in the future of the forecast (\textit{prediction window -- PW}). 

As case studies, we consider the time series made of multiple telemetry variables collected with IoT sensors from (1) an industrial wrapper machine, \textcolor{red}{(2) a blood refrigerator, and (3) a nitrogen generator}.

\textcolor{red}{In the three data sets, } IoT sensors collect physical quantities to monitor the system status. The emission of alert codes signals the occurrence of anomalous conditions.  The goal of the described data-driven predictive maintenance scenario is to anticipate the occurrence of a fault (i.e., a malfunctioning indicated by an alert code) within a future time interval (i.e., the prediction window) using historical data (i.e., the reading window). The presence of labeled data (i.e., the actual alert codes emitted by the machine in past work sessions) makes it possible to formulate the predictive maintenance problem as a binary classification task addressed with a fully supervised approach.

\textcolor{red}{Non-neural} Machine Learning (ML) and Deep Learning (DL) are popular methods for addressing predictive maintenance tasks. ML methods have been widely used in diverse applications \cite{Pertselakis2019, Khorsheed2020, Proto2019, Kaparthi2020, Leukel2022} and recent works also tested DL methods \cite{Alves2020, Dix2022}. A few works have compared supervised ML and DL methods for failure prediction in time series data \cite{Colone2019, Javeed2022, Dix2022} and evaluated the influence of either the reading window  (e.g., \cite{Kaparthi2020, Leahy2018, Proto2019}) or of the prediction window (e.g., \cite{Bonnevay2019, Colone2019, FigueroaBarraza2020, Khorsheed2020, Kusiak2012}). Only a few consider both parameters \cite{Li2014, Leukel2022}.

The contribution of the paper can be summarized as follows:
\begin{itemize}
    \item We compare three ML methods (Logistic Regression, Random Forest, and Support Vector Machine) \textcolor{red}{and three DL methods (LSTM, ConvLSTM, and Transformer) on three novel industrial data sets with multiple telemetry data}.
    \item We study the effect of varying the size of both the reading window and the prediction window in the context of failure prediction and discuss the consequences of these choices on the performances. 
    \item \textcolor{red}{We evaluate the diversity of patterns preceding faults using the Euclidean distance between time series  windows and the spectral entropy measured on whole data sets as a global complexity metrics, showing that DL approaches outperform ML approaches significantly only for complex data sets with more diverse patterns. In contrast, for simpler datasets where patterns exhibit greater uniformity, both DL and ML approaches produce comparable results, with DL algorithms not introducing substantial improvements.}    
    \item \textcolor{red}{All methods lose predictive power when the horizon enlarges because the temporal correlation between the input and the predicted event tends to vanish.} 
    \item \textcolor{red}{When patterns are diverse, the amount of historical data becomes more influential. In general, augmenting the amount of input is not always beneficial.}
    \item We publish the data sets and the compared algorithms to ensure reproducibility and allow the scientific community to extend the comparison to further ML and DL methods\footnote{Available at \url{https://github.com/nicolopinci/polimi_failure_prediction} (as of November 2023)}. 
\end{itemize}

The remainder of this paper is structured as follows: Section \ref{sec:related_work} surveys the related work, Section \ref{sec:method} presents the methodology adopted for pre-processing the data and making predictions, Section \ref{sec:results} presents and discusses the results, and Section \ref{sec:conclusions} presents the conclusions and proposes future work.

\section{Related work}
\label{sec:related_work}

Anomaly detection and failure prediction are two related fields applied to diverse data, including time series \cite{Laptev2015, Ren2019, Chen2019, Leukel2021, Zangrando2023}, and images \cite{Carrera2017, Si2018, Bionda2022}. This research focuses on failure prediction applied to multivariate time series data acquired by multiple IoT sensors connected to an industrial machine. \textcolor{red}{Several works have surveyed failure prediction on time series \cite{Xue2007, Leukel2021, Ramezani2021}.}

Failure prediction approaches belong to three categories, depending on the availability and use of data labels: supervised \cite{Alves2020, Leukel2022}, semi-supervised \cite{arxivSSL} and unsupervised \cite{Zhao2020}.  
Predicting failures in a future interval given labeled past data can be considered a supervised binary classification task \cite{Leukel2022}. The data set used in this research is labeled, making supervised methods a viable option.

\textcolor{red}{ML and DL} are two popular approaches for failure prediction in time series data. For supervised failure prediction, ML methods include decision trees (DT) \cite{Pertselakis2019, Khorsheed2020}, Gradient Boosting (GB) \cite{Proto2019, Khorsheed2020}, random forests (RF) \cite{Pertselakis2019, Proto2019, Kaparthi2020, Khorsheed2020}, support vector machines (SVM) \cite{Pertselakis2019, Khorsheed2020}, and Logistic Regression (LR) \cite{Kaparthi2020, Leukel2022}. DL methods include neural network architectures such as recurrent neural networks (RNN) \cite{Dix2022} and convolutional neural networks (CNN) \cite{Dix2022}. Some works have compared supervised ML and DL methods for failure prediction in time series data \cite{Colone2019, Javeed2022, Dix2022}.

RF is one of the most used approaches \cite{Kusiak2012, nowaczyk2013towards, Prytz2015, Canizo2017, Leahy2018, Xiang2018, Mishra2018, Kulkarni2018, Pertselakis2019, Chen2019, Bonnevay2019, Proto2019, Kaparthi2020, Khorsheed2020, Leukel2022}. The work in \cite{Kusiak2012} compares RF with ML and DL approaches, including Neural Networks and SVM, and shows the superiority of RF in terms of accuracy for a prediction window not longer than 300 seconds for a case study of wind turbines. However, accuracy does not consider class unbalance and is not appropriate for evaluating the performance of failure prediction algorithms \cite{Leukel2021}. The work in \cite{Canizo2017} also considers the case of wind turbines and presents only the performances of RF using accuracy, sensitivity, and specificity for a prediction window of 1 hour. The work in \cite{Leahy2018} also uses RF on wind turbine data and measures performances using precision and recall for varying the reading windows up to 42 hours, showing rather poor performances. Other applications of RF in predictive maintenance range from refrigerator systems \cite{Kulkarni2018} to components of vending machines \cite{Xiang2018}, vehicles \cite{Chen2019}, and chemical processes \cite{Proto2019}.

SVM is another approach commonly used in supervised failure prediction \cite{Kusiak2012, Li2014, Susto2015, Xiang2018, Pertselakis2019, hamaide2019predictive, Orr2020, Khorsheed2020, Leukel2022}. The work in \cite{Li2014} presents an application of SVM to railway systems and evaluates the performances using False Positive Rate (FPR) and recall. It is one of the few approaches considering variation in both PW and RW. However, the assessment of the contribution of each window is not performed because they were changed together. The work in \cite{Khorsheed2020} compares ML algorithms (DT, RF, GB, and SVM) and shows that GB outperforms the other approaches, followed by RF. The work in \cite{Pertselakis2019} compares the accuracy of six ML algorithms on industrial data, including SVM, RF, and DT. It shows that DT not only surpasses RF but is also more interpretable. However, precision and recall would be necessary to evaluate the quality of predictions, as the data set is likely unbalanced. The work in \cite{Kaparthi2020} also measures performances using accuracy (for a balanced data set), compares RF, LR, and Ctree, and shows that (1) RF outperforms the other algorithms, and (2) the reading window size does not have a significant impact on the performances.

A few works contrast ML and  DL predictors. The work in \cite{Colone2019} compares the performances using AUC and shows that a small fully-connected neural network outperforms Na\"ive Bayes. However, other popular methods (e.g., SVM, LR, LSTM) are not evaluated. The work in \cite{Dix2022} is one of the few approaches considering a comparison between two DL approaches (a CNN and LSTM) and shows (1) the superiority of the convolutional network in terms of accuracy and (2) the decrease in accuracy as the prediction window increases. However, additional metrics would be necessary to evaluate performances on a likely unbalanced data set. 

\textcolor{red}{Considering alternative DL approaches, aside from the commonly employed LSTM-based architectures \cite{Yu2019, Fagerstrm2019, Aung2023}, there is noteworthy research interest on ConvLSTM models. The work in \cite{Jin2023} finds that ConvLSTM exhibits comparable performance to Random Forest (RF) when predicting  faults in wind turbine gearbox systems. The work in \cite{Alos2022} shows the effectiveness of a ConvLSTM-based architecture for failure detection. Both works do not  analyze the impact of the size of RW and PW nor compare alternative DL architectures. ConvLSTM has also proved effective in forecasting the value of telemetry variables,  a  preliminary step for fault prediction in diverse systems \cite{https://doi.org/10.48550/arxiv.1506.04214, Szarek2023, Wu20221}. } \textcolor{red}{Transformer-based architectures  have been applied to  anomaly detection \cite{https://doi.org/10.48550/arxiv.2201.07284, Huang2020}, and fault diagnosis \cite{Jin2022, Wu2022}. The work in \cite{Gao2023} shows the effectiveness of a Transformer-based architecture for the fault prediction task in an Electric Power Communication Network. In contrast to LSTM-based architectures, Transformers can capture long-range dependencies and offer parallel processing capabilities, which makes them suitable also for time series tasks  over long intervals.}

Several works investigate the failure prediction task by forecasting the occurrence of a  failure within a pre-determined time frame \cite{Li2014, Proto2019, Colone2019, Kaparthi2020}.  Some studies evaluate the impact of either the reading window length \cite{Li2014, Proto2019, Kaparthi2020} or the prediction window length \cite{Colone2019, Bonnevay2019, FigueroaBarraza2020}. Only a few \cite{Li2014, Leukel2022} evaluate the impact of varying the reading and the prediction window. Considering the surveyed approaches, the work in \cite{Leukel2022} is the only one that proposes a comparison between different ML algorithms varying both the RW and PW and assessing the contribution of each. However, it does not consider DL algorithms in the comparison.

In summary, a few works on supervised failure prediction for multivariate time series evaluate the variation of both RW and PW. However, none compares ML and DL algorithms in this respect. Moreover, none of the surveyed papers considers the presence of discrete sessions in the telemetry time series, \textcolor{red}{which is instead the case of the wrapping machine data set}. This research investigates the contribution of both the RW and PW variations and compares diverse ML and DL approaches \textcolor{red}{on three novel industrial data sets featuring both discrete session-based and continuous time machines.}

\section{Method}
\label{sec:method}
This section outlines the experimental design employed to evaluate the impact of the reading/prediction windows and the hyperparameter selection on the performance of ML and DL approaches. It describes the data sets used and explains the data pre-processing procedures,  the experimental methodology, and the performance metrics. 

\subsection{Data sets}
\label{subsec:data set}

\subsubsection{Wrapping machine}

\textcolor{red}{Wrapping machines are  systems used for packaging and comprise various components such as motors, sensors, and controllers. The monitored exemplars are semi-automatic machines that wrap objects carried on pallets with stretch film. Each machine comprises: the \textit{turntable}, a central circular platform for  loading objects; the \textit{lifter},  a moving part that contains the wrapping film reel; the \textit{tower}, a column on which the lifter moves; the \textit{platform motor}, a gear motor controlling the movements of the turntable with a chain drive, and the  \textit{lifting motor}, an electric motor for lifting the carriage.
}

\textcolor{red}{The functioning of the machine is cyclic. A cycle  consists of five steps:
\begin{inparaenum}[1)]
\item Loading the products on a pallet at the center of the rotating platform. 
\item Tying the film's trailing end to the pallet's base.
\item  Starting the wrapping cycle, executed according to previously set parameters. 
\item
 Manually cutting the film (through the cutter or with an external tool) and making it adhere to the wrapped products.
\item Removing the wrapped objects.    
\end{inparaenum}
}

The data set contains records from sensors installed on a wrapping machine collected over one year (01-06-2021 to 31-05-2022). 
The machine is equipped with sensors attached to different components, such as the rotating platform and the electric motors. An acquisition system gathers data from sensors and sends them to a storage server. 
Sensors measure over 100 quantities (e.g., temperatures, motor frequencies, and platform speeds). Each data item records a timestamp, and the last value measured by a sensor. Data are transmitted only when at least one sensor registers a variation for saving energy, resulting in variable data acquisition frequency. Data are grouped into work sessions that represent periods in which the machine is operating. Each work session has a starting point, manually set by the operator, corresponding to the start of an interval during which the machine wraps at least one pallet. A work session ends when the machine finishes wrapping the pallets or when a fatal alert occurs. Session-level metrics are collected, too, such as the number of completed pallets and the quantity of consumed film.
The data set contains information about alerts thrown by the machine during the power-on state (i.e., when the machine is powered on, which does not necessarily mean it is producing pallets). An alert is a numerical code that refers to a specific problem that impacts the functioning of the machine.

Depending on the changes detected by sensors, the irregular sampling time is addressed by resampling the time series with a frequency of 5 seconds, which gives a good trade-off between memory occupation and signal resolution. Missing values are filled in using the last available observation because the absence of new measurements indicates no change in the recorded value.

\subsubsection{Blood refrigerator}

\textcolor{red}{Blood refrigerators are systems designed for the safe storage of blood and its derivatives at specific temperatures. Each unit includes different components. \textit{Compressor}: pumps the refrigerant, increasing its pressure and temperature. \textit{Condenser}: cools down and condenses the refrigerant. \textit{Evaporator}: causes the refrigerant to evaporate for cooling the inside of the unit. \textit{Expansion Valve}: regulates the flow of refrigerant into the evaporator. \textit{Interconnecting Tubing}: facilitates the movement of the refrigerant between the compressor, condenser, expansion valve and evaporator.
}

\textcolor{red}{
A refrigeration cycle comprises the following phases.
\begin{inparaenum}[1)]
    \item Compression: The cycle begins with the compressor drawing in low-pressure, low-temperature refrigerant gas. The compressor then compresses the gas, which raises its pressure and temperature.
    \item Condensation: The high-pressure, high-tempera\-ture gas exits the compressor and enters the condenser coils. As the gas flows through these coils, it releases heat,  cools down, and condenses into a high-pressure liquid.
    \item Expansion: This high-pressure liquid then flows through the capillary tube or expansion valve. As it does, its pressure drops, causing a significant drop in temperature. The refrigerant exits this stage as a low-pressure, cool liquid.
    \item Evaporation: The low-pressure, cool liquid refrigerant enters the evaporator coils inside the refrigerator. Here, it absorbs heat from the interior, causing it to evaporate and turn back into a low-pressure gas. This process cools the interior of the refrigerator.
    \item Return to Compressor: The low-pressure gas then returns to the compressor and the cycle starts over.
\end{inparaenum}
}

\textcolor{red}{The data set includes information from IoT sensors that measure 58 variables (e.g., the status of the door of the refrigerator, its temperature, energy consumption, etc.). The series comprises data from 31-10-2022 to 30-12-2022 and records  only when a variable changes value. This results in an irregular sampling period varying from a few seconds to one hour. For this reason, we resampled the data set using the median of the original sampling frequency (i.e., $\approx 34$ seconds).}

\subsubsection{Nitrogen generator}

\textcolor{red}{Nitrogen generators  separate the nitrogen from the other gasses, such as the oxygen, in compressed air. They include: an \textit{Air compressor},  to supply the compressed air;  \textit{Carbon Molecular Sieves (CMSs)}, to filter other gasses from the nitrogen; \textit{ Absorption vessels},  to take the nitrogen not filtered out by the CMS; \textit{ Towers}, to increase the nitrogen production. A tower comprises one CMS, two or more absorption vessels and the space where the division between nitrogen and oxygen takes place; \textit{ Valves}, to direct the flow of air and nitrogen and regulate the absorption of the nitrogen by the vessels; a \textit{ Buffer Tank}, to store the purified nitrogen.
}

\textcolor{red}{
The generator works as follows:
\begin{inparaenum}[1)]    
    \item The air compressor   compresses the air to a high pressure and supplies it to the machine. 
    \item In the towers  the CMS adsorbs smaller gas molecules such as oxygen while allowing larger nitrogen molecules to pass through the sieve and go into the vessels. 
    \item The buffer tank receives the nitrogen gas from the vessels through the valves.
    \item The valves reduce the pressure in the current working tower to release the residual gasses, while in the other towers the pressure is increased to restart the process.
\end{inparaenum}
}

\textcolor{red}{The data set includes information from IoT sensors installed in the generator. A set of  64 variables measure the essential work parameters (e.g., the nitrogen pressure, CMS air pressure, etc.). The series comprises data from 01-08-2023 to 29-09-2023 and contains values recorded when a change in a variable occurs. This results in an irregular sampling period varying from a few seconds to four hours. For this reason, we resampled the data set using 1 minute as frequency.}

\subsection{Data Processing}
\label{sec:data_processing}

\textcolor{red}{Data processing is an essential step in the analysis of each dataset, involving the selection of pertinent variables associated with physical quantities and of the resampling period. 
To  characterize the diversity of the data sets, the anomalous patterns preceding faults can be considered. To this end, the   data set is first normalized using min-max normalization and then a set of fixed-size windows preceding each fault is defined. For every pair of such windows, denoted as $w_i$ and $w_j$, the Euclidean distance, $d_{ijkv}$, is calculated between pairs of data points at corresponding positions, indexed by $k$, for each variable, represented as $v$. The distance is computed between the points $p_{ikv}$ and $p_{jkv}$. The diversity of patterns is quantified as the mean value scaled in the range $[0,1]$:}

\begin{equation}
\label{eq:diversity}
    m = mean_{ijkv}(d_{ijkv})
\end{equation}

\textcolor{red}{To characterize the complexity of a data set as a whole, we compute the mean of the spectral entropies of the considered time series \cite{Tang2015}, as implemented in the \texttt{antropy} library\footnote{\url{https://raphaelvallat.com/antropy/build/html/generated/antropy.spectral_entropy.html} (as of November 2023).}, normalized between 0 and 1 and using the fft method \cite{Inouye1991}. Higher entropy values correspond to a higher time series complexity.}

\subsubsection{Wrapping machine}
The first phase is feature selection. Since not all the quantities are equally relevant, only 13 are kept based on the data description provided by the manufacturer and are summarized in Table \ref{tab:variables}. Most removed features were redundant or had constant values. For example, such features as the employed recipe and the firmware version are excluded because they do not describe the system dynamics, whereas such features as the variation of a variable value during the session are eliminated because they can be derived from the initial and final values.

\begin{table}[H]
\centering
\caption{The 13 variables obtained after pre-processing. The minimum and maximum values are reported for each variable and the measurement unit. The "Movement" column indicates whether the variable is used to determine whether any motors are moving.}
\label{tab:variables}
\resizebox{0.8\textwidth}{!}{%
\begin{tabular}{@{}ccccc@{}}
\toprule
\textbf{Variable name} & \textbf{Minimum} & \textbf{Maximum} & \textbf{Unit} & \textbf{Movement} \\ \midrule
\textbf{Platform Motor frequency} & 0 & 5200 & Hz & \checkmark \\
\textbf{Current speed cart} & 0 & 100 & \% & \checkmark \\
\textbf{Platform motor speed} & 0 & 100 & \% & \checkmark \\
\textbf{Lifting motor speed} & 0 & 88 & RPM & \checkmark \\
\textbf{Platform rotation speed} & 0 & 184 & RPM & \checkmark \\
\textbf{Slave rotation speed} & 0 & 184 & m/min & \checkmark \\
\textbf{Lifting speed rotation} & 0 & 73 & m/min & \checkmark \\
\textbf{Flag roping} & 0 & 31 & $\mu m$ & \\
\textbf{Platform Position} & 0 & 359 & ° & \\
\textbf{Temperature hoist drive} & 0 & 55 & °C & \\
\textbf{Temperature platform drive} & 0 & 61 & °C & \\
\textbf{Temperature slave drive} & 0 & 55 & °C & \\
\textbf{Total film tension} & -100 & 9900 & \% & \\ \bottomrule
\end{tabular}%
}
\end{table}

\begin{figure}
  \centering
  \includegraphics[width=0.7\textwidth]{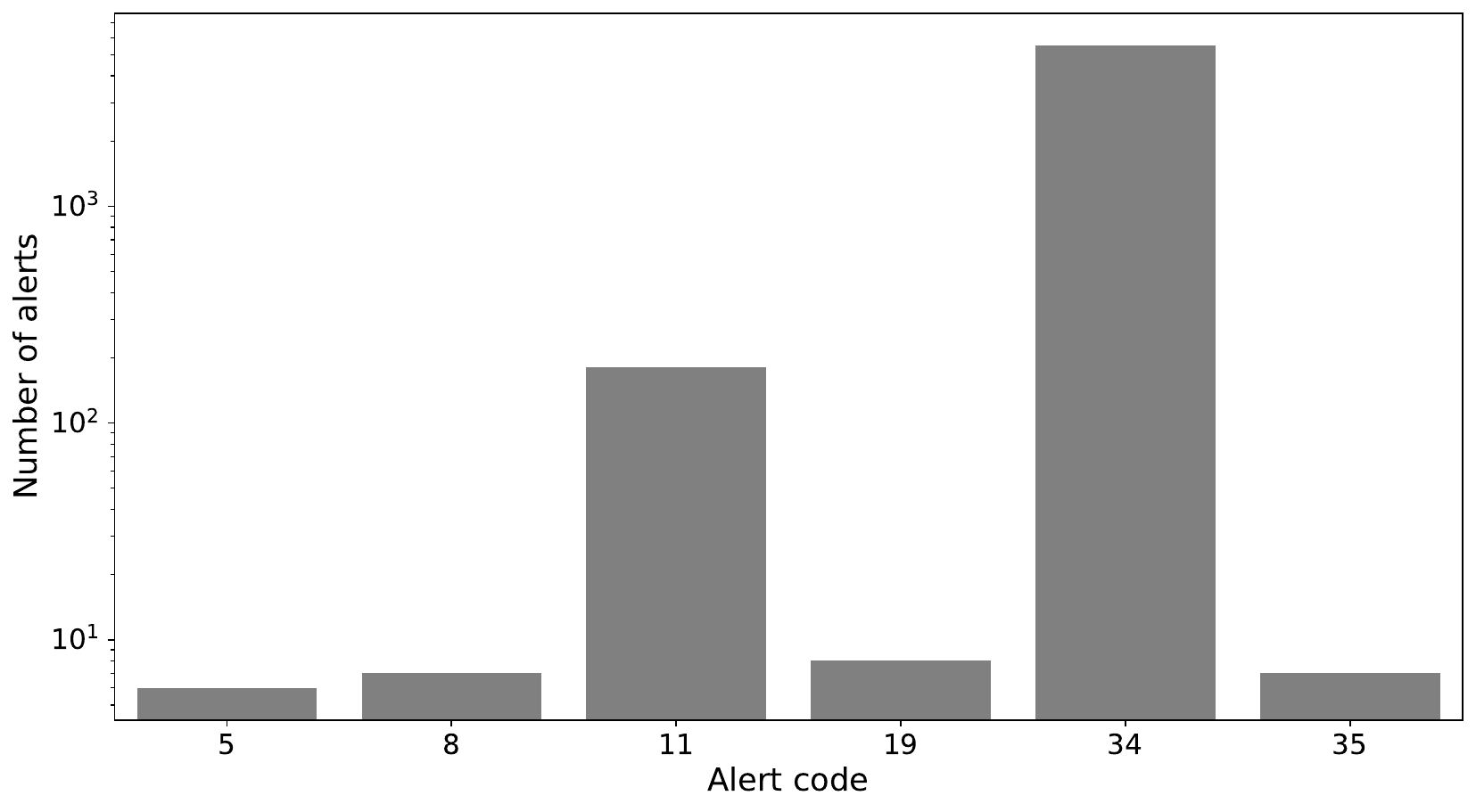}
     \caption{Distribution of the alert codes on a logarithmic scale. Alert 34 is the most frequent but must be discarded because it is unreliable. Alert 11 is the second most frequent. The remaining alerts have less than ten occurrences in the entire data set.}
  \label{fig:errors_per_machine}
\end{figure}

The second phase is selecting relevant alert codes for the anticipation of machine malfunction. Figure \ref{fig:errors_per_machine} presents the distribution of the alert codes most relevant according to the manufacturer. Most of them are rare (i.e., are observed less than 10 times), and alerts 11 (platform motor inverter protection) and 34 (machine in emergency condition) are the most common ones. Alert 34 is thrown when the operator presses an emergency button. However, this occurrence heavily depends on human behavior because the operators often use the emergency button as a quick way to turn the machine off, as confirmed by the plant manager. For this reason, alert 34 is discarded, and the prediction task focuses only on alert 11. \textcolor{red}{This alert is fundamental for the correct functioning of the machine, because the inverter controls the frequency or power supplied to the motor and thus controls its rotation speed. The improper  control of the rotation speed hinders the wrapping operation.}

\begin{figure}
  \centering
  \begin{subfigure}{0.48\textwidth}
    \includegraphics[width=\linewidth]{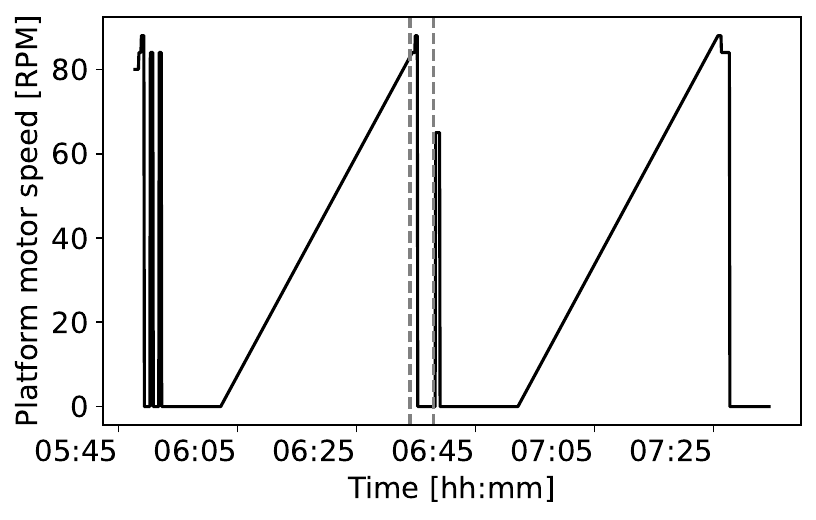}
  \end{subfigure}
  \begin{subfigure}{0.48\textwidth}
    \includegraphics[width=\linewidth]{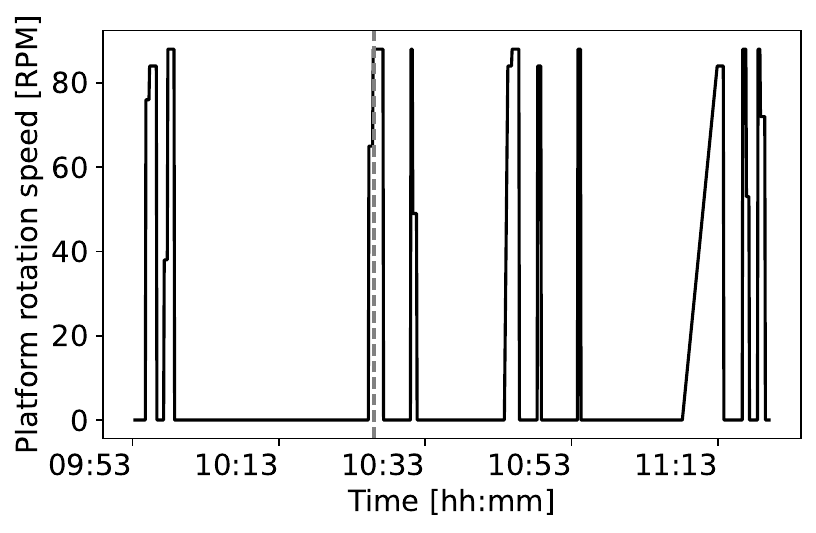}
  \end{subfigure}
  
  \begin{subfigure}{0.48\textwidth}
    \includegraphics[width=\linewidth]{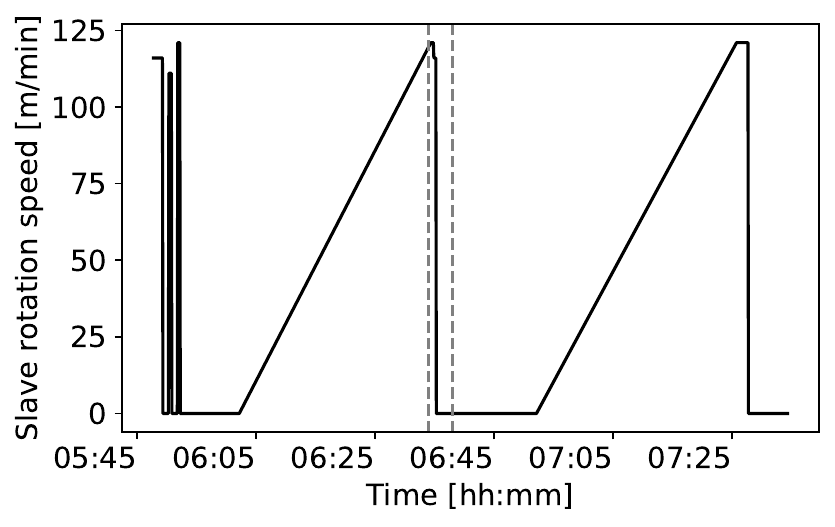}
  \end{subfigure}
  \begin{subfigure}{0.48\textwidth}
    \includegraphics[width=\linewidth]{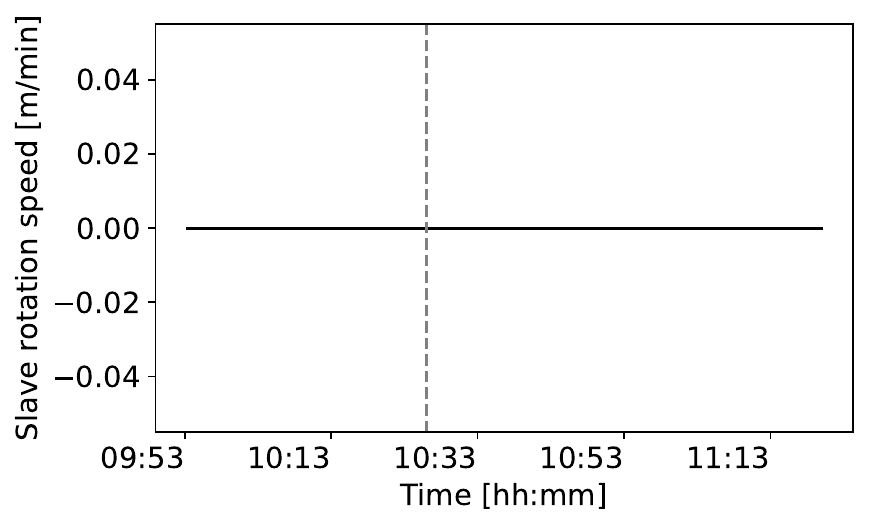}
  \end{subfigure}
  
  \begin{subfigure}{0.48\textwidth}
    \includegraphics[width=\linewidth]{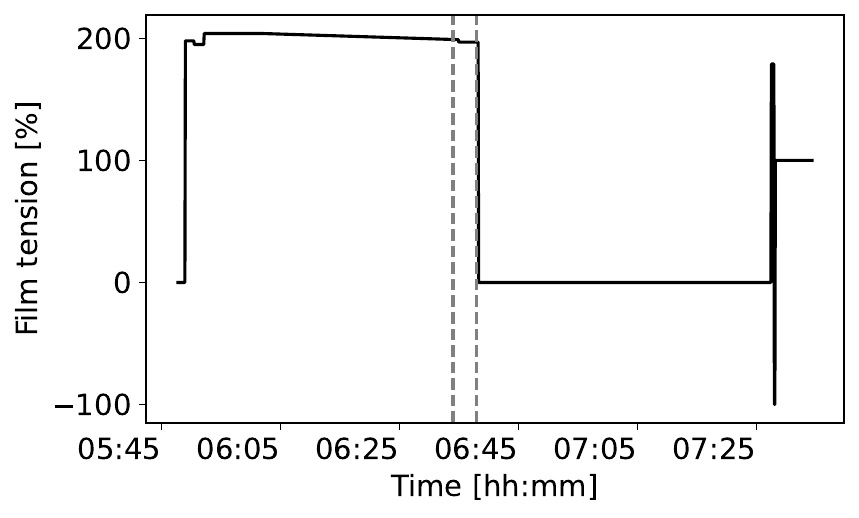}
  \end{subfigure}
  \begin{subfigure}{0.48\textwidth}
    \includegraphics[width=\linewidth]{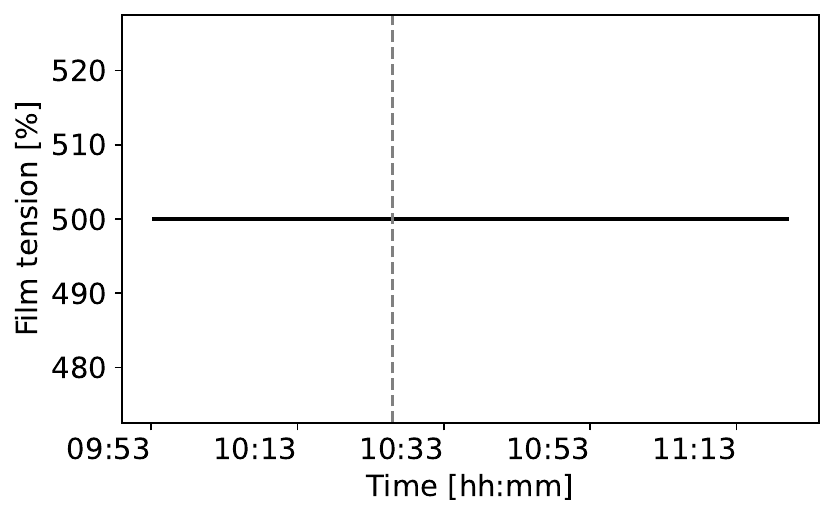}
  \end{subfigure}

  \caption{\textcolor{red}{Examples of diverse patterns preceding faults in the wrapping machine data set. The patterns involve three variables and the faults are displayed as dashed vertical lines. The example on the left shows two faults observed on 15-06-2021. They are preceded by a slow decrease in the film tension and after the second fault the slave motor has a speed of zero. On the right, the fault observed on 30-06-2021 is preceded by a sudden acceleration of the platform rotation speed, while the slave motor does not move, but the film tension remains constant.}}
    \label{fig:wrapping_patterns}
\end{figure}

\textcolor{red}{In the examined data set, alert 11 is preceded by diverse time-dependent patterns, and the mean Euclidean distance defined in Equation \ref{eq:diversity} is $\approx 0.26$ for a window of 15 minutes. Figure \ref{fig:wrapping_patterns} shows two examples of patterns preceding faults. The spectral entropy is $\approx 0.72$.}

The work session boundaries provided in the data set also depend on human intervention. Some inconsistent session-level metrics and a large number of sessions that do not produce any pallets are observed. A more reliable definition of a work session can be inferred from the telemetry variables. Intuitively, a session start time is introduced when at least one motion variable increases its value, and a session end time is created when all motion variables have decreased their value in the last ten minutes. Algorithm \ref{algo:session_definition} specifies how sessions are defined.

\begin{algorithm}
\caption{Computation of session boundaries}\label{algo:session_definition}
\begin{algorithmic}[1] 
\REQUIRE $ds$, the data set containing, for each timestamp, an array with the values of the movement variables, structured as a hash map with the timestamp $ts$ as the key and the movement variables array $mv$ as the value
\ENSURE $sessions$, a map that associates a session number to each timestamp and -1 to out-of-session timestamps 

\STATE $latest \leftarrow \textsc{null}$
\STATE $flag \leftarrow \FALSE$
\STATE $sessions \leftarrow HashMap()$
\STATE $counter \leftarrow 1$

\FOR{$ts \in ds.timestamps$}
  \STATE $sessions[ts] \leftarrow -1$
  \IF{$ts \neq min(ds.timestamps)$}
    \STATE $pts \leftarrow max(\{x: x \in ds.timestamps \; \AND \; x < ts\})$
    \STATE $\delta \leftarrow ds[ts].mv - ds[pts].mv$

    \IF{$\exists \, e \in \delta \: | \: e > 0 $}
      \STATE $latest \leftarrow ts$
      \STATE $flag \leftarrow$ \TRUE
      \STATE $sessions[ts] \leftarrow counter$
    \ELSE
      \IF{$flag =$ \TRUE}
        \STATE $sessions[ts] \leftarrow counter$
        \IF{$ts - latest > 10 $ minutes}
          \STATE $counter \leftarrow counter + 1$
          \STATE $flag \leftarrow$ \FALSE
        \ENDIF
      \ENDIF
    \ENDIF
  \ELSE
    \STATE $latest \leftarrow ts$
  \ENDIF
\ENDFOR
\end{algorithmic}
\end{algorithm}

\begin{figure}
  \centering
  \includegraphics[width=\textwidth]{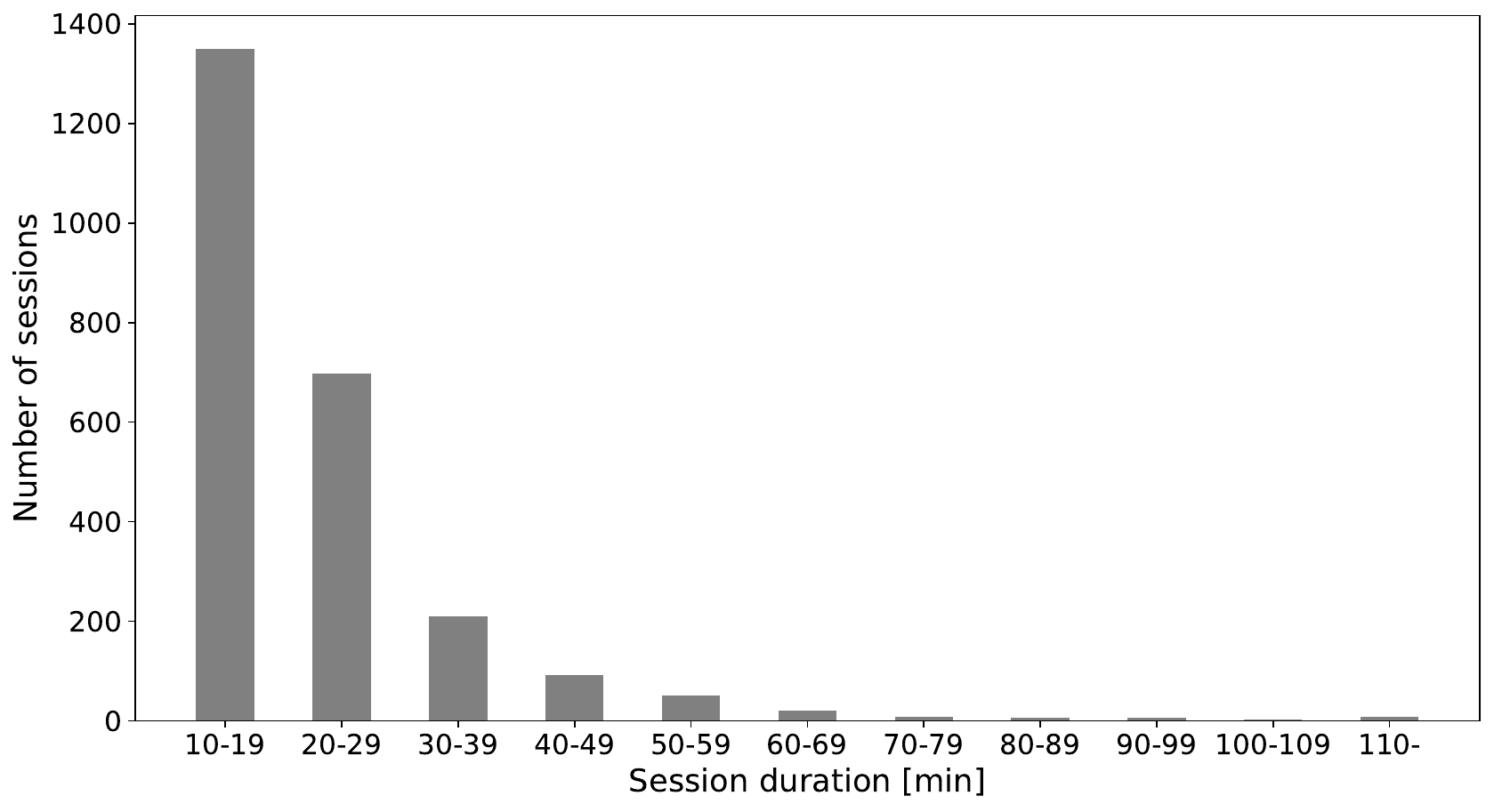}
  \caption{Distribution of the duration of work sessions. }
  \label{fig:new_sessions_duration}
\end{figure}

Figure \ref{fig:new_sessions_duration} shows the distribution of the work session duration computed from the telemetry data. The idle time intervals between sessions are neglected in further processing.

After pre-processing, it is possible to compute the distance of each data sample within a work session to the successive alert, i.e., the Time To Failure (TTF).

\subsubsection{Blood refrigerator}
\textcolor{red}{The data comprise variables directly measured by sensors (base variables) and variables derived from the base ones. For instance, the base variable "Door status" represents the status of the refrigerator door and has an associated  derived variable "Door opening daily counter", which counts every day when the door is open. We keep the 12 most significant  variables, shown in Table \ref{tab:variables_blood_refrigerators}}.

\begin{table}[H]
\caption{The 12 variables of the blood refrigerator}
\label{tab:variables_blood_refrigerators}
\centering
\resizebox{0.8\textwidth}{!}{%
\begin{tabular}{@{}ccccc@{}}
\toprule
\textbf{Variable Name}                        & \textbf{Minimum} & \textbf{Maximum} & \textbf{unit} \\ \midrule
\textbf{Product Temperature Base}    & -30.8         & 5.7           & °C            \\
\textbf{Evaporator temperature base} & -37.8         & 18.2          & °C            \\
\textbf{Condenser temperature base} & 15.8          & 36.7          & °C            \\
\textbf{Power supply}                 & 134.0            & 146.0            & V             \\
\textbf{Instant power consumption}  & 0.0              & 661.0            & W             \\
\textbf{Signal}                     & -113.0           & 85.0             & dBm           \\
\textbf{Door alert}                           & 0                & 1                &               \\
\textbf{Door close}                           & 0                & 1                &               \\
\textbf{Door open}                            & 0                & 1                &               \\
\textbf{Machine cooling}                      & 0                & 1                &               \\
\textbf{Machine defrost}                      & 0                & 1                &               \\
\textbf{Machine pause}                        & 0                & 1                &     \\ \midrule       
\end{tabular}%
}
\end{table}

\begin{figure}
    \centering

    \begin{subfigure}{0.48\textwidth}
        \includegraphics[width=\linewidth]{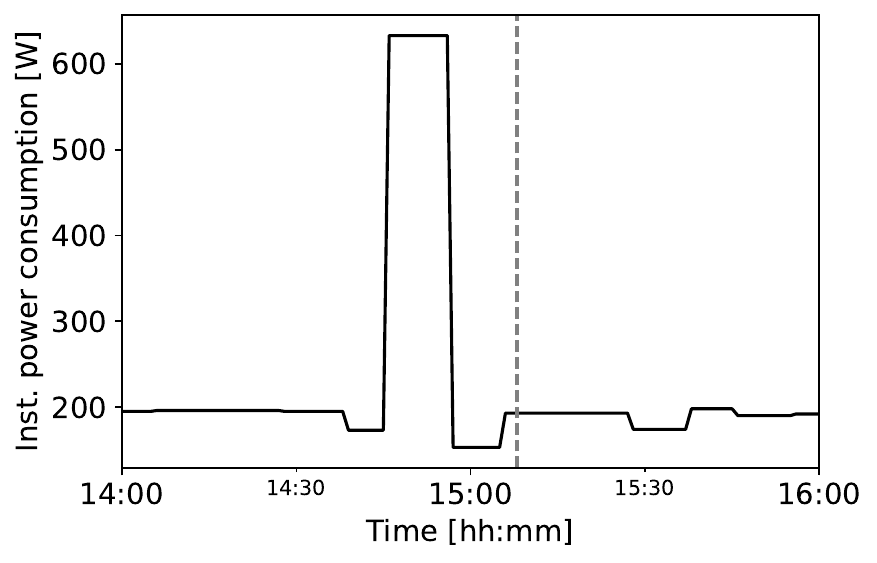}
        \caption{}
    \end{subfigure}
    \begin{subfigure}{0.48\textwidth}
        \includegraphics[width=\linewidth]{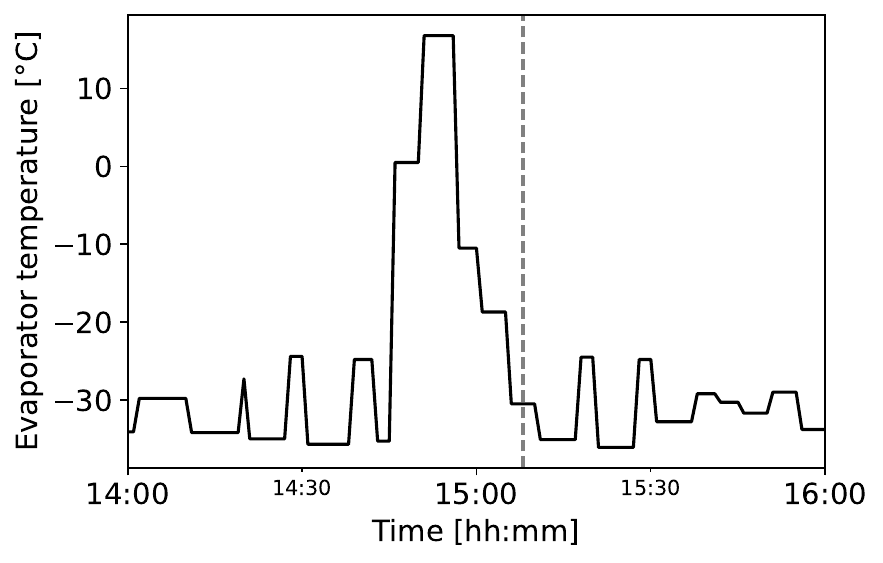}
        \caption{}
    \end{subfigure}
    
    \begin{subfigure}{0.48\textwidth}
        \includegraphics[width=\linewidth]{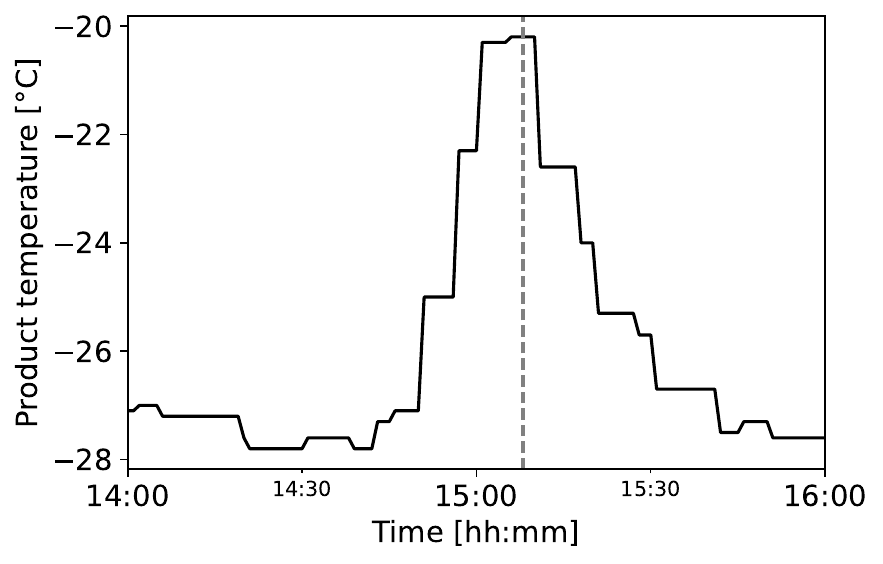}
        \caption{}
    \end{subfigure}

    \caption{\textcolor{red}{Examples of  patterns preceding a fault in a blood refrigerator, observed on 22-11-2022. Before the fault (shown by a dashed line), the instant power consumption increases and then decreases after a short time, the evaporator temperature decreases of $\approx 40 ^\circ C$ after a sudden increase, and the product temperature has an increase.}}
    \label{fig:blood_patterns}
\end{figure}

\textcolor{red}{The data set also contains  alarms (e.g., High-Temperature Compartment, Blackout, Defrost Timeout Failure, etc.). Specifically, the data set originally contained 15 types of alarms, but those described as critical by the manufacturer are the ones presented in Table \ref{tab:alarms_blood_refrigerators}. After an analysis of such two alarms, we have selected "alarm 5" as the  prediction target because "alarm 1" has too few occurrences, making the analysis of performances unfeasible. Storing blood at low temperature is crucial, as higher temperature can lead to bacterial growth \cite{Saxena1990}, hemolysis \cite{Blaine2018} and increased risk of adverse reactions, including life-threatening conditions \cite{aalaei2014blood}. Figure \ref{fig:blood_patterns} presents the typical anomalous pattern preceding a fault for three variables. In this case, the mean Euclidean distance defined in Equation \ref{eq:diversity} is $\approx 0.21$ for a window of 15 minutes, $\approx 20\%$ less than the one of the wrapping machine data set. The spectral entropy is $\approx 0.59$, indicating a lower complexity with respect to the wrapper machines data set.}

\begin{table}[H]
\caption{The alarms of the blood refrigerator}
\label{tab:alarms_blood_refrigerators}
\centering
\resizebox{0.8\textwidth}{!}{%
\begin{tabular}{@{}ccccc@{}}
\toprule
\textbf{Name   of the alarm} & \textbf{Description}         & \textbf{Number   of data points} \\ \midrule
\textbf{alarm 1}             & High-Temperature Compartment & 2                                \\
\textbf{alarm 5}             & High Product Temperature     & 188        \\ \midrule
\end{tabular}%
}
\end{table}

\subsubsection{Nitrogen generator}
\textcolor{red}{
The data are collected by IoT sensors over two months (01-08-2023 to 29-09-2023). We have considered only the variables that are directly measured from the sensors and discarded the derived ones.  Table \ref{tab:variables_nitrogen} shows the relevant variables, which record the changes of the physical status of the system.}

\begin{figure}
    \centering

    \begin{subfigure}{0.48\textwidth}
        \includegraphics[width=\linewidth]{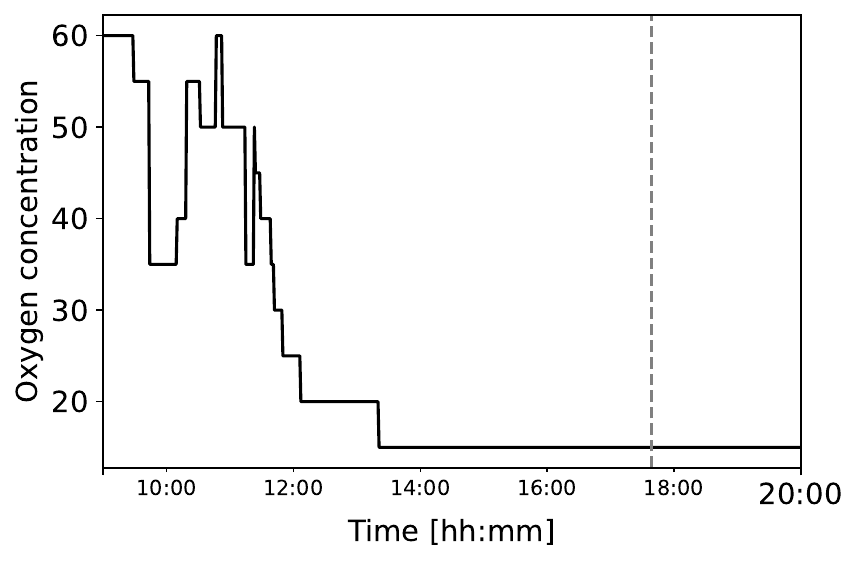}
        \caption{}
    \end{subfigure}
    \begin{subfigure}{0.48\textwidth}
        \includegraphics[width=\linewidth]{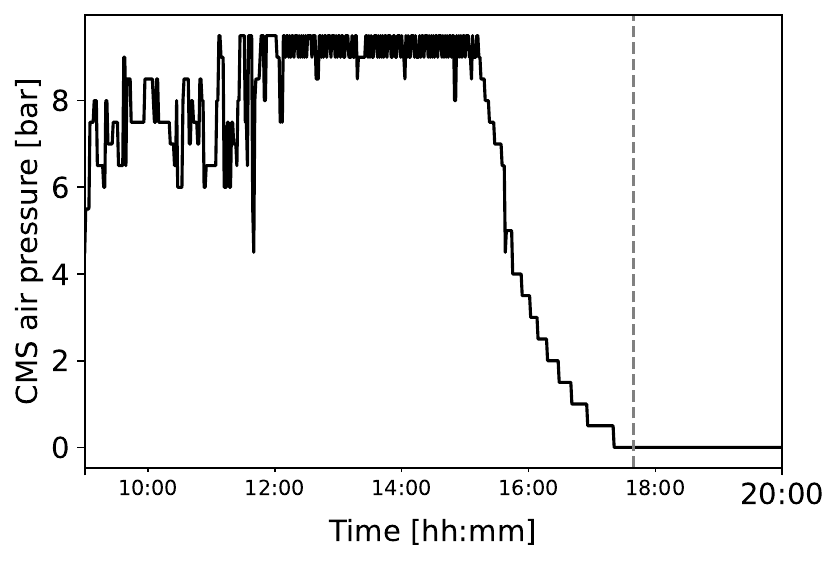}
        \caption{}
    \end{subfigure}
    
    \begin{subfigure}{0.48\textwidth}
        \includegraphics[width=\linewidth]{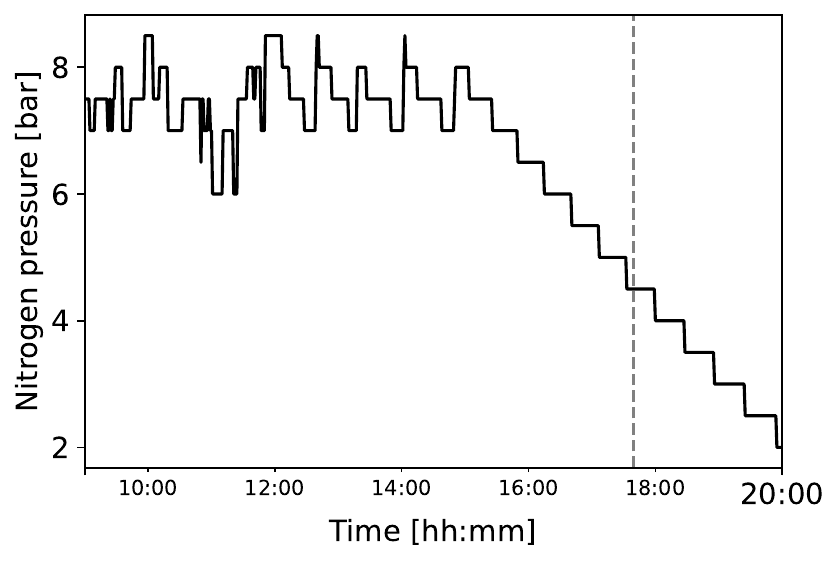}
        \caption{}
    \end{subfigure}

    \caption{\textcolor{red}{A typical pattern in a nitrogen generator, observed on 01-08-2023. In this case, before a fault (indicated with a gray dashed line), the oxygen concentration drops, the CMS air pressure decreases of of $\approx 8$ bar, and the nitrogen pressure decreases of $\approx 4$ bar, making the generation of nitrogen impossible.}}
    \label{fig:nitrogen_patterns}
\end{figure}

\begin{table}[H]
\caption{The 4 variables of the nitrogen generator}
\label{tab:variables_nitrogen}
\centering
\resizebox{0.8\textwidth}{!}{%
\begin{tabular}{@{}cccc@{}}
\toprule
{\color[HTML]{000000} \textbf{Variable Name}} & \textbf{Minimum} & \textbf{Maximum} & \textbf{unit} \\ \midrule
\textbf{CMS air pressure} & 0.0 & 9.5 & bar \\
\textbf{Oxygen base concentration} & 5.0 & 500.0 &  \\
\textbf{Nitrogen pressure} & 0.0 & 9.0 & bar \\
\textbf{Oxygen over threshold} & 0 & 1 & 
\\  \midrule
\end{tabular}%
}
\end{table}
\textcolor{red}{The nitrogen generator can produce different  alarms, including "Air pressure too high", and "Oxygen failure - Second threshold reached". However, only one type of alarm is present in the data set: "CMS pressurization fail", whose frequency is shown in Table \ref{tab:alarms_nitrogen}. A fail in CMS pressurization can hinder the separation of oxygen and nitrogen, as mentioned in \cite{LlosaTanco2016}. Figure \ref{fig:nitrogen_patterns} presents the typical anomalous pattern of three variables preceding a fault. The mean Euclidean distance defined in Equation \ref{eq:diversity} is $\approx 0.09$ for a window of 15 minutes, $\approx 65\%$ less than the one of the wrapping machine data set, and $\approx 57\%$ less than the one of the blood refrigerator data set. The spectral entropy is $\approx 0.53$, indicating a lower complexity with respect to both the wrapper machines and the blood refrigerator data sets.}

\begin{table}[H]
\caption{The alarm of the nitrogen generator}
\centering
\label{tab:alarms_nitrogen}
\resizebox{0.8\textwidth}{!}{%
\begin{tabular}{@{}ccc@{}}
\toprule
\textbf{Name of the alarm} & \textbf{Description} & \textbf{Number of data points} \\ \midrule
\textbf{CMS pressurization fail} & Pressurization of nitrogen sieve failed & 54 \\ \midrule
\end{tabular}%
}
\end{table}

\subsection{Definition of the Reading and Prediction Windows}

The task for comparing ML and DL methods is predicting the occurrence of alerts within a future time interval (i.e., a prediction window PW) given historical data (i.e., a reading window RW). This task is formulated as a binary classification, which assigns a label (failure/no failure) to each RW \cite{Leukel2021}. \textcolor{red}{Failures were provided by the manufacturer, which has measured them using sensors installed in the considered machines.}
A failure label associated with an RW \textcolor{red}{is assigned when the corresponding PW contains at least an alert code. It }is interpreted as a high probability of an alert occurring in the PW next to it.
Given a session of length $N$ and an RW size $S_{RW}$, $N - S_{RW} + 1$ reading windows are extracted. Each RW starts at timestamp $t_i$, with $i = 1 \ldots N - S_{RW} + 1$.
For an RW $(t_j \ldots t_j + S_{RW})$, the corresponding PW starts at time stamp $t_j + S_{RW} +1$. \textcolor{red}{The proposed approach can be applied in absence of sessions without loss of generality, because the entire interval of the data set can be considered as  one session.}

\subsection{Class Unbalance}
Alerts are anomalies and thus, by definition, rarer than normal behaviors \cite{Chandola2009}. For this reason, the number of RWs with label \textit{no failure} is significantly greater than that of RWs with label \textit{failure}, leading to a heavily imbalanced data set. 
\textcolor{red}{For all the data sets,} Random UnderSampling (RUS) \cite{Prusa2015, Zuech2021} is applied to balance the class distribution by making the cardinality of the majority class comparable to that of the minority class. The algorithm randomly selects and removes observations from the majority class until it achieves the desired equilibrium between the two classes. \textcolor{red}{In the case of the wrapping machine,} RUS is applied separately on each train set (comprising 4 folds) and test set (1 fold) for each combination of RW and PW sizes. To prevent the presence of similar data in the train and test sets (i.e., partially overlapping data), after the definition of the test set, the partially overlapping windows in the train set are neglected by RUS.
The RUS step is applied 10 times, as proposed in \cite{Leukel2022}. \textcolor{red}{In the blood refrigerator and nitrogen generator data sets, RUS is applied on the train and validation sets during the training phase, but not on the test set, where the test $F_1$ score is evaluated assigning a different weight to the \textit{failure} and \textit{no failure} classes in order to account for the unbalance.}

\subsection{Algorithms and hyperparameter tuning}

\textcolor{red}{The algorithms for which the influence of the RW and PW size is assessed include Logistic Regression (LR), as a representative of the Generalized Linear Models family \cite{STEFANSKI1986}, Random Forest (RF), as a representative of the Ensemble Methods family \cite{Breiman2001}, Support Vector Machine (SVM), as a representative of the Kernel Methods family \cite{SnchezA2003}, Long Short-Term Memory (LSTM) \cite{SiamiNamini2019}, Transformers \cite{Zeng2023} and ConvLSTM \cite{NIPS2015_07563a3f}, as representatives of DL methods.}

For SVM, RF, and LR, the implementation used in this work is from the Python library \texttt{scikit-learn}\footnote{\url{https://scikit-learn.org/stable/} (as of November 2023)}, while \textcolor{red}{DL} models use \texttt{TensorFlow}\footnote{\url{https://www.tensorflow.org/} (as of November 2023)}. \textcolor{red}{The tested network architectures of the DL models and the hyperparameters resulting from the hyperparameter search  of all models can be found in the project repository.}

For LR, $C$ is the inverse of the regularization strength, and smaller values indicate a stronger regularization.

In the case of RF, the search varies the number of internal estimators $N_E$ (i.e., the number of decision trees) and the maximum number of features that can be selected at each split (in this context, a feature is defined as the value of a variable at a specific timestamp). 

In the case of SVM, the $C$ coefficient is the regularization hyperparameter multiplied by the squared $L_2$ penalty and is inversely proportional to the strength of the regularization. Small values of $C$ lead to a low penalty for misclassified points, while if $C$ is large, SVM tries to minimize the number of misclassified examples.

\textcolor{red}{Considering LSTM, in the case of the wrapping machine} the only hyperparameters that vary are the type of LSTM layer, Unidirectional or Bidirectional, and the loss function.  For the latter, we compare the sigmoid$F_1$ loss with $\beta = 1$ and $\eta = 0$ (from now on referred to as $F_1$ loss) \cite{https://doi.org/10.48550/arxiv.2108.10566} and the Binary Cross-Entropy (BCE) loss. The use of BiLSTM has been shown to be effective in the works \cite{Abduljabbar2021, SiamiNamini2019}, which focus on forecasting problems. The two compared loss functions have been proven to deliver the top performances in \cite{https://doi.org/10.48550/arxiv.2108.10566}. Each loss function serves a distinct purpose and offers advantages and disadvantages. The $F_1$ loss considers precision and recall, thus providing a balanced model performance evaluation. In this case, minimizing the $F_1$ loss means maximizing the $F_1$ score for the "Failure" class (i.e., reducing the missed "Failure" occurrences). The BCE loss is a widely used loss function for binary classification tasks and is symmetric with respect to the two classes. It measures the dissimilarity between predicted probabilities and target labels, encouraging the model to improve its probability estimations. BCE loss encourages the model to output probabilities for each class, providing richer information about the model's confidence in its prediction. However, it does not inherently consider the interplay between precision and recall and is not the most suitable choice for tasks where a balance between precision and recall is essential.

\textcolor{red}{In the case of Transformers, the Binary Cross-Entropy (BCE) loss function is employed \cite{Tang2023}, and two  hyperparameters are relevant: the number of attention heads and the number of transformer blocks. The former affects the  capacity of the model to capture diverse patterns and relationships in the data, with  higher values improving the ability to discern patterns. The latter  increases the complexity of the model, thus allowing it to capture more complex patterns but also augmenting the risk of overfitting.}

\textcolor{red}{In the case of ConvLSTM \cite{NIPS2015_07563a3f}, BCE is  chosen as the loss function and the hyperparameter search (described in the project repository) has varied both the kernel size and the number of filters. Increasing the kernel size extends the receptive field  and augmenting the number of filters allows capturing more diverse and complex spatiotemporal features.}

\subsection{Training and evaluation}

Results are assessed using the macro $F_1$ score, which extends the $F_1$ score, defined in Equation \ref{eq:f1}. The $F_1$ score depends on precision and recall, defined respectively in Equation \ref{eq:precision} and Equation \ref{eq:recall}, where TP stands for the number of true positives, FP for the number of false positives, TN for the number of true negatives, and FN for the number of false negatives.

\begin{equation}
\label{eq:f1}
F_1 = 2\cdot \frac{PREC \cdot REC}{PREC + REC} 
\end{equation}

\begin{equation}
PREC = \frac{TP}{TP + FP}
\label{eq:precision}
\end{equation}

\begin{equation}
REC = \frac{TP}{TP + FN}
\label{eq:recall}
\end{equation}

The macro $F_1$ score is the average of the $F_1$ scores computed for each class independently. To compute the $F_1$ score for each class, the model treats that class as the positive class and the other class as the negative class. Let $F_{1,f}$ be the $F_1$ score computed when class "Failure" is the positive one and $F_{1,n}$ be the $F_1$ score computed when class "No failure" is the positive one. Then, the macro $F_1$ score is given by Equation \ref{eq:macro_f1}.

\begin{equation}
\label{eq:macro_f1}
MF_1 = \frac{F_{1,f} + F_{1,n}}{2}
\end{equation}

\textcolor{red}{The range of RW and PW sizes is adapted to the dynamics of the machines (a slow varying behavior requires testing less window sizes than a fast changing one). For the wrapping machine data set, we use 9 values for the PW (0.25, 0.5, 1, 1.5, 2, 2.5, 3, 3.5, and 4 hours) and 6 values for the RW (10, 15, 20, 25, 30, and 35 minutes). For the  blood refrigerator,  we use 4 values for the PW (0.5, 1, 1.5, and 2 hours) and 5 values for the RW (10, 15, 20, 25, and 30 minutes). For the nitrogen generator data set,  we use 6 values for the PW (0.5, 1, 1.5, 2, 3 and 5 hours) and 5 values for the RW (10, 15, 20, 25, and 30 minutes).  For all the data sets, the minimum PW size is set to the smallest time span necessary for avoiding an unrecoverable machine failure, so that the operator can intervene.}

\textcolor{red}{The validation procedure is also adapted to the characteristics of the different use cases. In  the wrapping machine data set there are only 13 alarms, which yield $\approx 500$ \textit{failure} RWs in the whole time series. Thus the number of failure RWs in the test set would be too small to test adequately  the performances.}
Thus we adopt a training and evaluation procedure based on k-fold cross-validation (with $k=5$). Each fold is identified by $f_k$ with $k \in [1, 5]$. The procedure also relies on 10 RUS instances $r_{ik}$ for the test set and 10 RUS instances $p_{ik}$ for the train set, 6 algorithms $a_j$, with $j \in \{$RF, LR, SVM, LSTM, \textcolor{red}{ConvLSTM, Transformer}$\}$, and the hyperparameter settings for each algorithm ($h_{jl}$). Each combination of RW and PW sizes is used to compare the models with a procedure consisting of six steps:

\begin{enumerate}
  \item For each k-fold split, the data are divided into a train set $TrS_k$, comprising the folds $f_m$ with $m \neq k$, and a test set $TeS_k$, corresponding to $f_k$.
  \item For each $TrS_k$ and $TeS_k$, RUS is applied to balance the classes,
  obtaining 10 RUS instances $r_{ik}$ for the test set and 10 RUS instances $p_{ik}$ for the train set.
  \item For each $p_{ik}$, the four presented algorithms $a_j$, with their hyperparameters $h_{jl}$, are trained on the train folds and evaluated on the RUS instance $r_{ik}$ on the test fold, obtaining $j\times l$ macro $F_1$ scores on the test set, denoted as $s_{ikjl}$.
  \item For each algorithm and hyperparameter setting, the mean of the results is computed as $m_{jl} = \text{mean}_{i,k}(s_{ikjl})$
  \item Then, the maximum macro $F_1$ score for each algorithm is computed as $b_j = \max_l(m_{jl})$
  \item Finally, the best macro $F_1$ score is computed as $B_s = \max_j(b_j)$, and the best algorithm as $B_a = \text{argmax}_j(b_j)$
\end{enumerate}

\textcolor{red}{For the blood refrigerator and nitrogen generator data sets, a higher number of \textit{failure} RWs is available and thus k-fold cross validation is unnecessary. Each data set is first divided into a train, validation and test set. The model is trained on the training set and the best combination of hyperparameters  is determined as the one with the best macro $F_1$ score on the validation set using Bayesian Optimization. Finally, the best configuration of each algorithm is evaluated on the test set.}

\section{Results}
\label{sec:results}
This section reports the macro $F_1$ scores (from now on referred to as $F_1$ scores) and the macro average precision and recall of the compared algorithms computed with the abovementioned procedure. It discusses the effect of RW and PW selection on SVM, RF, LR, LSTM, \textcolor{red}{ConvLSTM and Transformers} prediction performances.

\subsection{Wrapping machine}
\begin{figure}
    \centering
    \includegraphics[width=\textwidth]{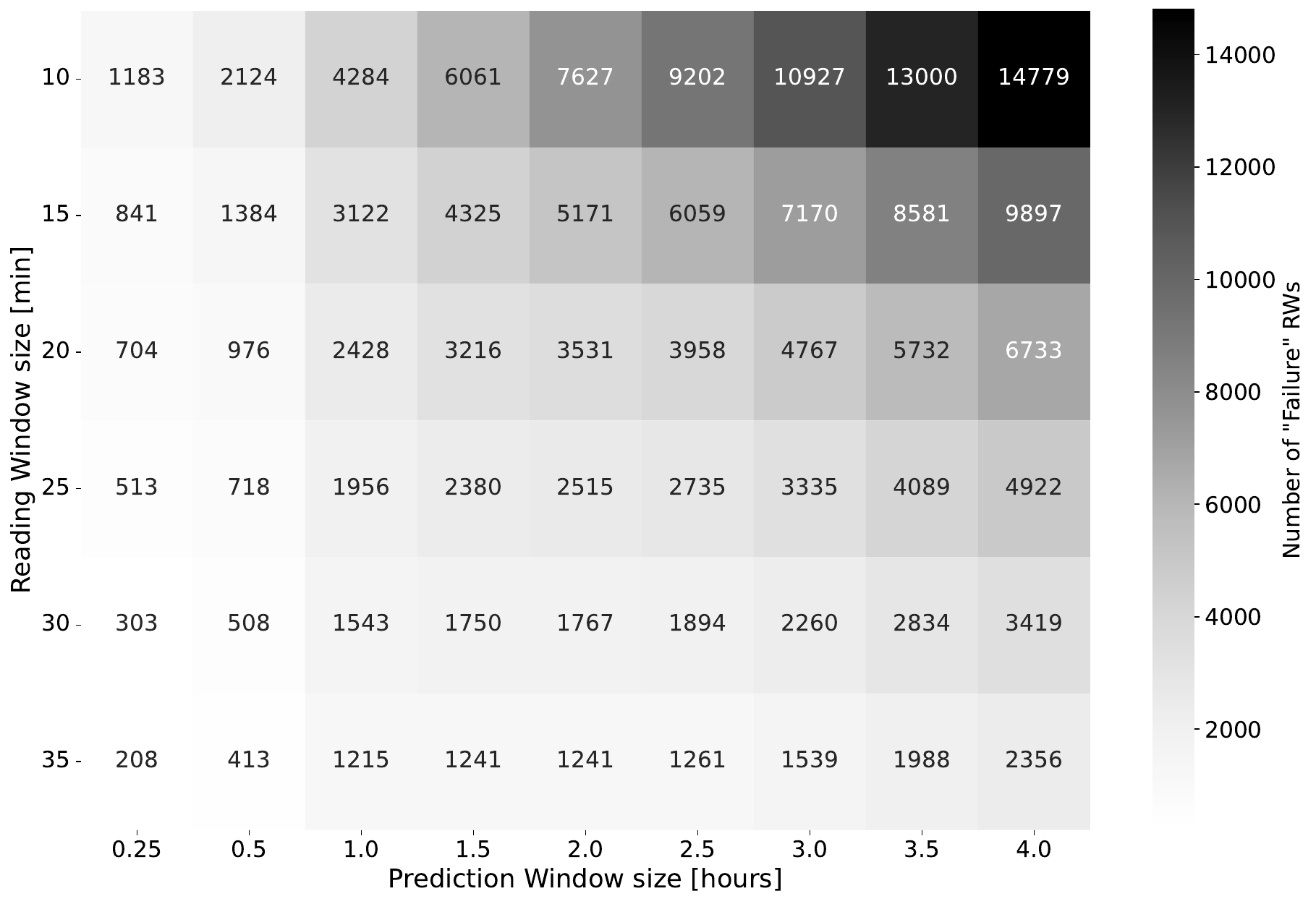}
    \caption{The number of RWs with class "Failure" as RW and PW sizes vary. The highest number of "Failure" labels is observed for short RWs and long PWs.}
    \label{fig:support}
\end{figure}
\begin{table}[]
\renewcommand{\arraystretch}{1.4}
\centering
\caption{Classification results on the $F_1$ metrics for the wrapping machine case study, varying both the reading and prediction window, for SVM, RF, LR, LSTM, \textcolor{red}{ConvLSTM, and Transformers}. Lighter background corresponds to better results. Empty cells indicate that the results cannot be computed due to the lack of "Failure" labels in some folds.}
\label{tab:bc_f1_global_results}
\resizebox{\textwidth}{!}{%
\begin{tabular}{|c|c|ccccccccc|}
\hline
 &  & \multicolumn{9}{c|}{\textbf{Prediction   Window (PW)}} \\ \cline{3-11} 
\multirow{-2}{*}{\textbf{Reading   Window (RW)}} & \multirow{-2}{*}{\textbf{Algorithm}} & \multicolumn{1}{c|}{\textbf{0.25h}} & \multicolumn{1}{c|}{\textbf{0.5h}} & \multicolumn{1}{c|}{\textbf{1.0h}} & \multicolumn{1}{c|}{\textbf{1.5h}} & \multicolumn{1}{c|}{\textbf{2.0h}} & \multicolumn{1}{c|}{\textbf{2.5h}} & \multicolumn{1}{c|}{\textbf{3.0h}} & \multicolumn{1}{c|}{\textbf{3.5h}} & \multicolumn{1}{c|}{\textbf{4.0h}} \\ \hline
 & \textbf{ConvLSTM} & \multicolumn{1}{l|}{\cellcolor[HTML]{AAAAAA}{\color[HTML]{F1F1F1} 0.684}} & \multicolumn{1}{l|}{\cellcolor[HTML]{AEAEAE}0.691} & \multicolumn{1}{l|}{\cellcolor[HTML]{777777}{\color[HTML]{F1F1F1} 0.577}} & \multicolumn{1}{l|}{\cellcolor[HTML]{6C6C6C}{\color[HTML]{F1F1F1} 0.554}} & \multicolumn{1}{l|}{\cellcolor[HTML]{626262}{\color[HTML]{F1F1F1} 0.533}} & \multicolumn{1}{l|}{\cellcolor[HTML]{5A5A5A}{\color[HTML]{F1F1F1} 0.517}} & \multicolumn{1}{l|}{\cellcolor[HTML]{5E5E5E}{\color[HTML]{F1F1F1} 0.524}} & \multicolumn{1}{l|}{\cellcolor[HTML]{555555}{\color[HTML]{F1F1F1} 0.506}} & \cellcolor[HTML]{4A4A4A}{\color[HTML]{F1F1F1} 0.483} \\ \cline{2-11} 
 & \textbf{LR} & \multicolumn{1}{l|}{\cellcolor[HTML]{848484}{\color[HTML]{F1F1F1} 0.604}} & \multicolumn{1}{l|}{\cellcolor[HTML]{818181}{\color[HTML]{F1F1F1} 0.598}} & \multicolumn{1}{l|}{\cellcolor[HTML]{6A6A6A}{\color[HTML]{F1F1F1} 0.549}} & \multicolumn{1}{l|}{\cellcolor[HTML]{5B5B5B}{\color[HTML]{F1F1F1} 0.518}} & \multicolumn{1}{l|}{\cellcolor[HTML]{545454}{\color[HTML]{F1F1F1} 0.503}} & \multicolumn{1}{l|}{\cellcolor[HTML]{515151}{\color[HTML]{F1F1F1} 0.498}} & \multicolumn{1}{l|}{\cellcolor[HTML]{616161}{\color[HTML]{F1F1F1} 0.532}} & \multicolumn{1}{l|}{\cellcolor[HTML]{595959}{\color[HTML]{F1F1F1} 0.514}} & \cellcolor[HTML]{5A5A5A}{\color[HTML]{F1F1F1} 0.516} \\ \cline{2-11} 
 & \textbf{LSTM} & \multicolumn{1}{l|}{\cellcolor[HTML]{CECECE}0.758} & \multicolumn{1}{l|}{\cellcolor[HTML]{C7C7C7}0.743} & \multicolumn{1}{l|}{\cellcolor[HTML]{929292}{\color[HTML]{F1F1F1} 0.634}} & \multicolumn{1}{l|}{\cellcolor[HTML]{828282}{\color[HTML]{F1F1F1} 0.599}} & \multicolumn{1}{l|}{\cellcolor[HTML]{7A7A7A}{\color[HTML]{F1F1F1} 0.583}} & \multicolumn{1}{l|}{\cellcolor[HTML]{636363}{\color[HTML]{F1F1F1} 0.536}} & \multicolumn{1}{l|}{\cellcolor[HTML]{6A6A6A}{\color[HTML]{F1F1F1} 0.550}} & \multicolumn{1}{l|}{\cellcolor[HTML]{656565}{\color[HTML]{F1F1F1} 0.540}} & \cellcolor[HTML]{585858}{\color[HTML]{F1F1F1} 0.512} \\ \cline{2-11} 
 & \textbf{RF} & \multicolumn{1}{l|}{\cellcolor[HTML]{777777}{\color[HTML]{F1F1F1} 0.577}} & \multicolumn{1}{l|}{\cellcolor[HTML]{747474}{\color[HTML]{F1F1F1} 0.571}} & \multicolumn{1}{l|}{\cellcolor[HTML]{303030}{\color[HTML]{F1F1F1} 0.429}} & \multicolumn{1}{l|}{\cellcolor[HTML]{2A2A2A}{\color[HTML]{F1F1F1} 0.417}} & \multicolumn{1}{l|}{\cellcolor[HTML]{232323}{\color[HTML]{F1F1F1} 0.402}} & \multicolumn{1}{l|}{\cellcolor[HTML]{1C1C1C}{\color[HTML]{F1F1F1} 0.388}} & \multicolumn{1}{l|}{\cellcolor[HTML]{191919}{\color[HTML]{F1F1F1} 0.382}} & \multicolumn{1}{l|}{\cellcolor[HTML]{191919}{\color[HTML]{F1F1F1} 0.381}} & \cellcolor[HTML]{1E1E1E}{\color[HTML]{F1F1F1} 0.391} \\ \cline{2-11} 
 & \textbf{SVM} & \multicolumn{1}{l|}{\cellcolor[HTML]{B1B1B1}0.697} & \multicolumn{1}{l|}{\cellcolor[HTML]{9C9C9C}{\color[HTML]{F1F1F1} 0.654}} & \multicolumn{1}{l|}{\cellcolor[HTML]{7D7D7D}{\color[HTML]{F1F1F1} 0.590}} & \multicolumn{1}{l|}{\cellcolor[HTML]{676767}{\color[HTML]{F1F1F1} 0.543}} & \multicolumn{1}{l|}{\cellcolor[HTML]{5C5C5C}{\color[HTML]{F1F1F1} 0.521}} & \multicolumn{1}{l|}{\cellcolor[HTML]{5A5A5A}{\color[HTML]{F1F1F1} 0.517}} & \multicolumn{1}{l|}{\cellcolor[HTML]{5C5C5C}{\color[HTML]{F1F1F1} 0.520}} & \multicolumn{1}{l|}{\cellcolor[HTML]{4B4B4B}{\color[HTML]{F1F1F1} 0.485}} & \cellcolor[HTML]{474747}{\color[HTML]{F1F1F1} 0.476} \\ \cline{2-11} 
\multirow{-6}{*}{\textbf{10   minutes}} & \textbf{Transformer} & \multicolumn{1}{l|}{\cellcolor[HTML]{B9B9B9}0.714} & \multicolumn{1}{l|}{\cellcolor[HTML]{989898}{\color[HTML]{F1F1F1} 0.646}} & \multicolumn{1}{l|}{\cellcolor[HTML]{7B7B7B}{\color[HTML]{F1F1F1} 0.585}} & \multicolumn{1}{l|}{\cellcolor[HTML]{7A7A7A}{\color[HTML]{F1F1F1} 0.584}} & \multicolumn{1}{l|}{\cellcolor[HTML]{656565}{\color[HTML]{F1F1F1} 0.540}} & \multicolumn{1}{l|}{\cellcolor[HTML]{4B4B4B}{\color[HTML]{F1F1F1} 0.486}} & \multicolumn{1}{l|}{\cellcolor[HTML]{515151}{\color[HTML]{F1F1F1} 0.498}} & \multicolumn{1}{l|}{\cellcolor[HTML]{494949}{\color[HTML]{F1F1F1} 0.482}} & \cellcolor[HTML]{3B3B3B}{\color[HTML]{F1F1F1} 0.452} \\ \hline
 & \textbf{ConvLSTM} & \multicolumn{1}{l|}{\cellcolor[HTML]{DCDCDC}0.787} & \multicolumn{1}{l|}{\cellcolor[HTML]{AAAAAA}{\color[HTML]{F1F1F1} 0.683}} & \multicolumn{1}{l|}{\cellcolor[HTML]{6C6C6C}{\color[HTML]{F1F1F1} 0.554}} & \multicolumn{1}{l|}{\cellcolor[HTML]{616161}{\color[HTML]{F1F1F1} 0.530}} & \multicolumn{1}{l|}{\cellcolor[HTML]{696969}{\color[HTML]{F1F1F1} 0.547}} & \multicolumn{1}{l|}{\cellcolor[HTML]{616161}{\color[HTML]{F1F1F1} 0.532}} & \multicolumn{1}{l|}{\cellcolor[HTML]{6E6E6E}{\color[HTML]{F1F1F1} 0.558}} & \multicolumn{1}{l|}{\cellcolor[HTML]{595959}{\color[HTML]{F1F1F1} 0.514}} & \cellcolor[HTML]{515151}{\color[HTML]{F1F1F1} 0.498} \\ \cline{2-11} 
 & \textbf{LR} & \multicolumn{1}{l|}{\cellcolor[HTML]{C6C6C6}0.742} & \multicolumn{1}{l|}{\cellcolor[HTML]{696969}{\color[HTML]{F1F1F1} 0.548}} & \multicolumn{1}{l|}{\cellcolor[HTML]{5A5A5A}{\color[HTML]{F1F1F1} 0.516}} & \multicolumn{1}{l|}{\cellcolor[HTML]{616161}{\color[HTML]{F1F1F1} 0.530}} & \multicolumn{1}{l|}{\cellcolor[HTML]{585858}{\color[HTML]{F1F1F1} 0.513}} & \multicolumn{1}{l|}{\cellcolor[HTML]{4A4A4A}{\color[HTML]{F1F1F1} 0.483}} & \multicolumn{1}{l|}{\cellcolor[HTML]{5F5F5F}{\color[HTML]{F1F1F1} 0.526}} & \multicolumn{1}{l|}{\cellcolor[HTML]{656565}{\color[HTML]{F1F1F1} 0.539}} & \cellcolor[HTML]{656565}{\color[HTML]{F1F1F1} 0.540} \\ \cline{2-11} 
 & \textbf{LSTM} & \multicolumn{1}{l|}{\cellcolor[HTML]{F8F8F8}0.846} & \multicolumn{1}{l|}{\cellcolor[HTML]{CCCCCC}0.753} & \multicolumn{1}{l|}{\cellcolor[HTML]{A4A4A4}{\color[HTML]{F1F1F1} 0.671}} & \multicolumn{1}{l|}{\cellcolor[HTML]{7D7D7D}{\color[HTML]{F1F1F1} 0.590}} & \multicolumn{1}{l|}{\cellcolor[HTML]{888888}{\color[HTML]{F1F1F1} 0.612}} & \multicolumn{1}{l|}{\cellcolor[HTML]{7F7F7F}{\color[HTML]{F1F1F1} 0.593}} & \multicolumn{1}{l|}{\cellcolor[HTML]{7F7F7F}{\color[HTML]{F1F1F1} 0.593}} & \multicolumn{1}{l|}{\cellcolor[HTML]{696969}{\color[HTML]{F1F1F1} 0.547}} & \cellcolor[HTML]{707070}{\color[HTML]{F1F1F1} 0.562} \\ \cline{2-11} 
 & \textbf{RF} & \multicolumn{1}{l|}{\cellcolor[HTML]{717171}{\color[HTML]{F1F1F1} 0.565}} & \multicolumn{1}{l|}{\cellcolor[HTML]{1E1E1E}{\color[HTML]{F1F1F1} 0.391}} & \multicolumn{1}{l|}{\cellcolor[HTML]{272727}{\color[HTML]{F1F1F1} 0.410}} & \multicolumn{1}{l|}{\cellcolor[HTML]{101010}{\color[HTML]{F1F1F1} 0.363}} & \multicolumn{1}{l|}{\cellcolor[HTML]{0F0F0F}{\color[HTML]{F1F1F1} 0.361}} & \multicolumn{1}{l|}{\cellcolor[HTML]{171717}{\color[HTML]{F1F1F1} 0.376}} & \multicolumn{1}{l|}{\cellcolor[HTML]{111111}{\color[HTML]{F1F1F1} 0.365}} & \multicolumn{1}{l|}{\cellcolor[HTML]{0A0A0A}{\color[HTML]{F1F1F1} 0.349}} & \cellcolor[HTML]{121212}{\color[HTML]{F1F1F1} 0.367} \\ \cline{2-11} 
 & \textbf{SVM} & \multicolumn{1}{l|}{\cellcolor[HTML]{CCCCCC}0.753} & \multicolumn{1}{l|}{\cellcolor[HTML]{9E9E9E}{\color[HTML]{F1F1F1} 0.659}} & \multicolumn{1}{l|}{\cellcolor[HTML]{616161}{\color[HTML]{F1F1F1} 0.532}} & \multicolumn{1}{l|}{\cellcolor[HTML]{5E5E5E}{\color[HTML]{F1F1F1} 0.524}} & \multicolumn{1}{l|}{\cellcolor[HTML]{585858}{\color[HTML]{F1F1F1} 0.513}} & \multicolumn{1}{l|}{\cellcolor[HTML]{5B5B5B}{\color[HTML]{F1F1F1} 0.518}} & \multicolumn{1}{l|}{\cellcolor[HTML]{555555}{\color[HTML]{F1F1F1} 0.505}} & \multicolumn{1}{l|}{\cellcolor[HTML]{444444}{\color[HTML]{F1F1F1} 0.471}} & \cellcolor[HTML]{474747}{\color[HTML]{F1F1F1} 0.477} \\ \cline{2-11} 
\multirow{-6}{*}{\textbf{15   minutes}} & \textbf{Transformer} & \multicolumn{1}{l|}{\cellcolor[HTML]{EBEBEB}0.818} & \multicolumn{1}{l|}{\cellcolor[HTML]{A7A7A7}{\color[HTML]{F1F1F1} 0.677}} & \multicolumn{1}{l|}{\cellcolor[HTML]{888888}{\color[HTML]{F1F1F1} 0.613}} & \multicolumn{1}{l|}{\cellcolor[HTML]{727272}{\color[HTML]{F1F1F1} 0.567}} & \multicolumn{1}{l|}{\cellcolor[HTML]{707070}{\color[HTML]{F1F1F1} 0.563}} & \multicolumn{1}{l|}{\cellcolor[HTML]{555555}{\color[HTML]{F1F1F1} 0.506}} & \multicolumn{1}{l|}{\cellcolor[HTML]{585858}{\color[HTML]{F1F1F1} 0.513}} & \multicolumn{1}{l|}{\cellcolor[HTML]{545454}{\color[HTML]{F1F1F1} 0.503}} & \cellcolor[HTML]{474747}{\color[HTML]{F1F1F1} 0.477} \\ \hline
 & \textbf{ConvLSTM} & \multicolumn{1}{l|}{\cellcolor[HTML]{E6E6E6}0.807} & \multicolumn{1}{l|}{\cellcolor[HTML]{B2B2B2}0.699} & \multicolumn{1}{l|}{\cellcolor[HTML]{727272}{\color[HTML]{F1F1F1} 0.567}} & \multicolumn{1}{l|}{\cellcolor[HTML]{6E6E6E}{\color[HTML]{F1F1F1} 0.559}} & \multicolumn{1}{l|}{\cellcolor[HTML]{616161}{\color[HTML]{F1F1F1} 0.531}} & \multicolumn{1}{l|}{\cellcolor[HTML]{686868}{\color[HTML]{F1F1F1} 0.546}} & \multicolumn{1}{l|}{\cellcolor[HTML]{6B6B6B}{\color[HTML]{F1F1F1} 0.551}} & \multicolumn{1}{l|}{\cellcolor[HTML]{646464}{\color[HTML]{F1F1F1} 0.537}} & \cellcolor[HTML]{4D4D4D}{\color[HTML]{F1F1F1} 0.489} \\ \cline{2-11} 
 & \textbf{LR} & \multicolumn{1}{l|}{\cellcolor[HTML]{DDDDDD}0.789} & \multicolumn{1}{l|}{\cellcolor[HTML]{B0B0B0}0.696} & \multicolumn{1}{l|}{\cellcolor[HTML]{7C7C7C}{\color[HTML]{F1F1F1} 0.587}} & \multicolumn{1}{l|}{\cellcolor[HTML]{6C6C6C}{\color[HTML]{F1F1F1} 0.554}} & \multicolumn{1}{l|}{\cellcolor[HTML]{5A5A5A}{\color[HTML]{F1F1F1} 0.517}} & \multicolumn{1}{l|}{\cellcolor[HTML]{5B5B5B}{\color[HTML]{F1F1F1} 0.518}} & \multicolumn{1}{l|}{\cellcolor[HTML]{6D6D6D}{\color[HTML]{F1F1F1} 0.555}} & \multicolumn{1}{l|}{\cellcolor[HTML]{7B7B7B}{\color[HTML]{F1F1F1} 0.585}} & \cellcolor[HTML]{707070}{\color[HTML]{F1F1F1} 0.563} \\ \cline{2-11} 
 & \textbf{LSTM} & \multicolumn{1}{l|}{\cellcolor[HTML]{FFFFFF}0.861} & \multicolumn{1}{l|}{\cellcolor[HTML]{E1E1E1}0.798} & \multicolumn{1}{l|}{\cellcolor[HTML]{A1A1A1}{\color[HTML]{F1F1F1} 0.665}} & \multicolumn{1}{l|}{\cellcolor[HTML]{898989}{\color[HTML]{F1F1F1} 0.614}} & \multicolumn{1}{l|}{\cellcolor[HTML]{6E6E6E}{\color[HTML]{F1F1F1} 0.559}} & \multicolumn{1}{l|}{\cellcolor[HTML]{828282}{\color[HTML]{F1F1F1} 0.599}} & \multicolumn{1}{l|}{\cellcolor[HTML]{878787}{\color[HTML]{F1F1F1} 0.611}} & \multicolumn{1}{l|}{\cellcolor[HTML]{6D6D6D}{\color[HTML]{F1F1F1} 0.557}} & \cellcolor[HTML]{6D6D6D}{\color[HTML]{F1F1F1} 0.557} \\ \cline{2-11} 
 & \textbf{RF} & \multicolumn{1}{l|}{\cellcolor[HTML]{646464}{\color[HTML]{F1F1F1} 0.538}} & \multicolumn{1}{l|}{\cellcolor[HTML]{393939}{\color[HTML]{F1F1F1} 0.448}} & \multicolumn{1}{l|}{\cellcolor[HTML]{242424}{\color[HTML]{F1F1F1} 0.404}} & \multicolumn{1}{l|}{\cellcolor[HTML]{1F1F1F}{\color[HTML]{F1F1F1} 0.393}} & \multicolumn{1}{l|}{\cellcolor[HTML]{191919}{\color[HTML]{F1F1F1} 0.381}} & \multicolumn{1}{l|}{\cellcolor[HTML]{1D1D1D}{\color[HTML]{F1F1F1} 0.389}} & \multicolumn{1}{l|}{\cellcolor[HTML]{181818}{\color[HTML]{F1F1F1} 0.379}} & \multicolumn{1}{l|}{\cellcolor[HTML]{2A2A2A}{\color[HTML]{F1F1F1} 0.417}} & \cellcolor[HTML]{1A1A1A}{\color[HTML]{F1F1F1} 0.384} \\ \cline{2-11} 
 & \textbf{SVM} & \multicolumn{1}{l|}{\cellcolor[HTML]{DADADA}0.783} & \multicolumn{1}{l|}{\cellcolor[HTML]{737373}{\color[HTML]{F1F1F1} 0.568}} & \multicolumn{1}{l|}{\cellcolor[HTML]{525252}{\color[HTML]{F1F1F1} 0.499}} & \multicolumn{1}{l|}{\cellcolor[HTML]{434343}{\color[HTML]{F1F1F1} 0.468}} & \multicolumn{1}{l|}{\cellcolor[HTML]{4B4B4B}{\color[HTML]{F1F1F1} 0.486}} & \multicolumn{1}{l|}{\cellcolor[HTML]{4B4B4B}{\color[HTML]{F1F1F1} 0.485}} & \multicolumn{1}{l|}{\cellcolor[HTML]{494949}{\color[HTML]{F1F1F1} 0.481}} & \multicolumn{1}{l|}{\cellcolor[HTML]{373737}{\color[HTML]{F1F1F1} 0.443}} & \cellcolor[HTML]{434343}{\color[HTML]{F1F1F1} 0.468} \\ \cline{2-11} 
\multirow{-6}{*}{\textbf{20   minutes}} & \textbf{Transformer} & \multicolumn{1}{l|}{\cellcolor[HTML]{E2E2E2}0.799} & \multicolumn{1}{l|}{\cellcolor[HTML]{9A9A9A}{\color[HTML]{F1F1F1} 0.649}} & \multicolumn{1}{l|}{\cellcolor[HTML]{7A7A7A}{\color[HTML]{F1F1F1} 0.584}} & \multicolumn{1}{l|}{\cellcolor[HTML]{6E6E6E}{\color[HTML]{F1F1F1} 0.559}} & \multicolumn{1}{l|}{\cellcolor[HTML]{676767}{\color[HTML]{F1F1F1} 0.543}} & \multicolumn{1}{l|}{\cellcolor[HTML]{585858}{\color[HTML]{F1F1F1} 0.513}} & \multicolumn{1}{l|}{\cellcolor[HTML]{5A5A5A}{\color[HTML]{F1F1F1} 0.516}} & \multicolumn{1}{l|}{\cellcolor[HTML]{484848}{\color[HTML]{F1F1F1} 0.478}} & \cellcolor[HTML]{343434}{\color[HTML]{F1F1F1} 0.438} \\ \hline
 & \textbf{ConvLSTM} & \multicolumn{1}{l|}{\cellcolor[HTML]{C7C7C7}0.744} & \multicolumn{1}{l|}{\cellcolor[HTML]{939393}{\color[HTML]{F1F1F1} 0.635}} & \multicolumn{1}{l|}{\cellcolor[HTML]{5F5F5F}{\color[HTML]{F1F1F1} 0.527}} & \multicolumn{1}{l|}{\cellcolor[HTML]{666666}{\color[HTML]{F1F1F1} 0.541}} & \multicolumn{1}{l|}{\cellcolor[HTML]{585858}{\color[HTML]{F1F1F1} 0.512}} & \multicolumn{1}{l|}{\cellcolor[HTML]{6D6D6D}{\color[HTML]{F1F1F1} 0.557}} & \multicolumn{1}{l|}{\cellcolor[HTML]{5D5D5D}{\color[HTML]{F1F1F1} 0.523}} & \multicolumn{1}{l|}{\cellcolor[HTML]{6F6F6F}{\color[HTML]{F1F1F1} 0.561}} & \cellcolor[HTML]{5E5E5E}{\color[HTML]{F1F1F1} 0.524} \\ \cline{2-11} 
 & \textbf{LR} & \multicolumn{1}{l|}{\cellcolor[HTML]{C3C3C3}0.735} & \multicolumn{1}{l|}{\cellcolor[HTML]{8C8C8C}{\color[HTML]{F1F1F1} 0.620}} & \multicolumn{1}{l|}{\cellcolor[HTML]{6B6B6B}{\color[HTML]{F1F1F1} 0.552}} & \multicolumn{1}{l|}{\cellcolor[HTML]{616161}{\color[HTML]{F1F1F1} 0.531}} & \multicolumn{1}{l|}{\cellcolor[HTML]{5B5B5B}{\color[HTML]{F1F1F1} 0.519}} & \multicolumn{1}{l|}{\cellcolor[HTML]{616161}{\color[HTML]{F1F1F1} 0.530}} & \multicolumn{1}{l|}{\cellcolor[HTML]{797979}{\color[HTML]{F1F1F1} 0.580}} & \multicolumn{1}{l|}{\cellcolor[HTML]{868686}{\color[HTML]{F1F1F1} 0.608}} & \cellcolor[HTML]{767676}{\color[HTML]{F1F1F1} 0.575} \\ \cline{2-11} 
 & \textbf{LSTM} & \multicolumn{1}{l|}{\cellcolor[HTML]{DADADA}0.783} & \multicolumn{1}{l|}{\cellcolor[HTML]{ABABAB}{\color[HTML]{F1F1F1} 0.686}} & \multicolumn{1}{l|}{\cellcolor[HTML]{8D8D8D}{\color[HTML]{F1F1F1} 0.623}} & \multicolumn{1}{l|}{\cellcolor[HTML]{888888}{\color[HTML]{F1F1F1} 0.612}} & \multicolumn{1}{l|}{\cellcolor[HTML]{8B8B8B}{\color[HTML]{F1F1F1} 0.618}} & \multicolumn{1}{l|}{\cellcolor[HTML]{777777}{\color[HTML]{F1F1F1} 0.577}} & \multicolumn{1}{l|}{\cellcolor[HTML]{787878}{\color[HTML]{F1F1F1} 0.579}} & \multicolumn{1}{l|}{\cellcolor[HTML]{737373}{\color[HTML]{F1F1F1} 0.568}} & \cellcolor[HTML]{646464}{\color[HTML]{F1F1F1} 0.538} \\ \cline{2-11} 
 & \textbf{RF} & \multicolumn{1}{l|}{\cellcolor[HTML]{969696}{\color[HTML]{F1F1F1} 0.641}} & \multicolumn{1}{l|}{\cellcolor[HTML]{323232}{\color[HTML]{F1F1F1} 0.433}} & \multicolumn{1}{l|}{\cellcolor[HTML]{202020}{\color[HTML]{F1F1F1} 0.395}} & \multicolumn{1}{l|}{\cellcolor[HTML]{181818}{\color[HTML]{F1F1F1} 0.379}} & \multicolumn{1}{l|}{\cellcolor[HTML]{232323}{\color[HTML]{F1F1F1} 0.401}} & \multicolumn{1}{l|}{\cellcolor[HTML]{323232}{\color[HTML]{F1F1F1} 0.433}} & \multicolumn{1}{l|}{\cellcolor[HTML]{272727}{\color[HTML]{F1F1F1} 0.410}} & \multicolumn{1}{l|}{\cellcolor[HTML]{232323}{\color[HTML]{F1F1F1} 0.401}} & \cellcolor[HTML]{181818}{\color[HTML]{F1F1F1} 0.378} \\ \cline{2-11} 
 & \textbf{SVM} & \multicolumn{1}{l|}{\cellcolor[HTML]{888888}{\color[HTML]{F1F1F1} 0.613}} & \multicolumn{1}{l|}{\cellcolor[HTML]{585858}{\color[HTML]{F1F1F1} 0.512}} & \multicolumn{1}{l|}{\cellcolor[HTML]{333333}{\color[HTML]{F1F1F1} 0.436}} & \multicolumn{1}{l|}{\cellcolor[HTML]{272727}{\color[HTML]{F1F1F1} 0.411}} & \multicolumn{1}{l|}{\cellcolor[HTML]{2A2A2A}{\color[HTML]{F1F1F1} 0.416}} & \multicolumn{1}{l|}{\cellcolor[HTML]{333333}{\color[HTML]{F1F1F1} 0.436}} & \multicolumn{1}{l|}{\cellcolor[HTML]{494949}{\color[HTML]{F1F1F1} 0.482}} & \multicolumn{1}{l|}{\cellcolor[HTML]{282828}{\color[HTML]{F1F1F1} 0.413}} & \cellcolor[HTML]{333333}{\color[HTML]{F1F1F1} 0.435} \\ \cline{2-11} 
\multirow{-6}{*}{\textbf{25   minutes}} & \textbf{Transformer} & \multicolumn{1}{l|}{\cellcolor[HTML]{D5D5D5}0.772} & \multicolumn{1}{l|}{\cellcolor[HTML]{9F9F9F}{\color[HTML]{F1F1F1} 0.660}} & \multicolumn{1}{l|}{\cellcolor[HTML]{5A5A5A}{\color[HTML]{F1F1F1} 0.517}} & \multicolumn{1}{l|}{\cellcolor[HTML]{606060}{\color[HTML]{F1F1F1} 0.529}} & \multicolumn{1}{l|}{\cellcolor[HTML]{545454}{\color[HTML]{F1F1F1} 0.504}} & \multicolumn{1}{l|}{\cellcolor[HTML]{4F4F4F}{\color[HTML]{F1F1F1} 0.493}} & \multicolumn{1}{l|}{\cellcolor[HTML]{4E4E4E}{\color[HTML]{F1F1F1} 0.492}} & \multicolumn{1}{l|}{\cellcolor[HTML]{3A3A3A}{\color[HTML]{F1F1F1} 0.449}} & \cellcolor[HTML]{3A3A3A}{\color[HTML]{F1F1F1} 0.450} \\ \hline
 & \textbf{ConvLSTM} & \multicolumn{1}{l|}{{\color[HTML]{F1F1F1} }} & \multicolumn{1}{l|}{\cellcolor[HTML]{2F2F2F}{\color[HTML]{F1F1F1} 0.426}} & \multicolumn{1}{l|}{\cellcolor[HTML]{4D4D4D}{\color[HTML]{F1F1F1} 0.490}} & \multicolumn{1}{l|}{\cellcolor[HTML]{414141}{\color[HTML]{F1F1F1} 0.464}} & \multicolumn{1}{l|}{\cellcolor[HTML]{343434}{\color[HTML]{F1F1F1} 0.437}} & \multicolumn{1}{l|}{\cellcolor[HTML]{444444}{\color[HTML]{F1F1F1} 0.470}} & \multicolumn{1}{l|}{\cellcolor[HTML]{737373}{\color[HTML]{F1F1F1} 0.569}} & \multicolumn{1}{l|}{\cellcolor[HTML]{737373}{\color[HTML]{F1F1F1} 0.569}} & \cellcolor[HTML]{606060}{\color[HTML]{F1F1F1} 0.528} \\ \cline{2-2} \cline{4-11} 
 & \textbf{LR} & \multicolumn{1}{l|}{{\color[HTML]{F1F1F1} }} & \multicolumn{1}{l|}{\cellcolor[HTML]{3D3D3D}{\color[HTML]{F1F1F1} 0.456}} & \multicolumn{1}{l|}{\cellcolor[HTML]{575757}{\color[HTML]{F1F1F1} 0.510}} & \multicolumn{1}{l|}{\cellcolor[HTML]{464646}{\color[HTML]{F1F1F1} 0.474}} & \multicolumn{1}{l|}{\cellcolor[HTML]{444444}{\color[HTML]{F1F1F1} 0.471}} & \multicolumn{1}{l|}{\cellcolor[HTML]{505050}{\color[HTML]{F1F1F1} 0.495}} & \multicolumn{1}{l|}{\cellcolor[HTML]{666666}{\color[HTML]{F1F1F1} 0.542}} & \multicolumn{1}{l|}{\cellcolor[HTML]{767676}{\color[HTML]{F1F1F1} 0.574}} & \cellcolor[HTML]{616161}{\color[HTML]{F1F1F1} 0.532} \\ \cline{2-2} \cline{4-11} 
 & \textbf{LSTM} & \multicolumn{1}{l|}{{\color[HTML]{F1F1F1} }} & \multicolumn{1}{l|}{\cellcolor[HTML]{787878}{\color[HTML]{F1F1F1} 0.579}} & \multicolumn{1}{l|}{\cellcolor[HTML]{7C7C7C}{\color[HTML]{F1F1F1} 0.588}} & \multicolumn{1}{l|}{\cellcolor[HTML]{8A8A8A}{\color[HTML]{F1F1F1} 0.616}} & \multicolumn{1}{l|}{\cellcolor[HTML]{888888}{\color[HTML]{F1F1F1} 0.612}} & \multicolumn{1}{l|}{\cellcolor[HTML]{808080}{\color[HTML]{F1F1F1} 0.596}} & \multicolumn{1}{l|}{\cellcolor[HTML]{818181}{\color[HTML]{F1F1F1} 0.598}} & \multicolumn{1}{l|}{\cellcolor[HTML]{7B7B7B}{\color[HTML]{F1F1F1} 0.585}} & \cellcolor[HTML]{6C6C6C}{\color[HTML]{F1F1F1} 0.553} \\ \cline{2-2} \cline{4-11} 
 & \textbf{RF} & \multicolumn{1}{l|}{{\color[HTML]{F1F1F1} }} & \multicolumn{1}{l|}{\cellcolor[HTML]{121212}{\color[HTML]{F1F1F1} 0.367}} & \multicolumn{1}{l|}{\cellcolor[HTML]{0F0F0F}{\color[HTML]{F1F1F1} 0.360}} & \multicolumn{1}{l|}{\cellcolor[HTML]{0C0C0C}{\color[HTML]{F1F1F1} 0.353}} & \multicolumn{1}{l|}{\cellcolor[HTML]{0C0C0C}{\color[HTML]{F1F1F1} 0.353}} & \multicolumn{1}{l|}{\cellcolor[HTML]{0B0B0B}{\color[HTML]{F1F1F1} 0.352}} & \multicolumn{1}{l|}{\cellcolor[HTML]{0A0A0A}{\color[HTML]{F1F1F1} 0.349}} & \multicolumn{1}{l|}{\cellcolor[HTML]{0C0C0C}{\color[HTML]{F1F1F1} 0.353}} & \cellcolor[HTML]{090909}{\color[HTML]{F1F1F1} 0.347} \\ \cline{2-2} \cline{4-11} 
 & \textbf{SVM} & \multicolumn{1}{l|}{{\color[HTML]{F1F1F1} }} & \multicolumn{1}{l|}{\cellcolor[HTML]{010101}{\color[HTML]{F1F1F1} 0.332}} & \multicolumn{1}{l|}{\cellcolor[HTML]{090909}{\color[HTML]{F1F1F1} 0.348}} & \multicolumn{1}{l|}{\cellcolor[HTML]{050505}{\color[HTML]{F1F1F1} 0.339}} & \multicolumn{1}{l|}{\cellcolor[HTML]{040404}{\color[HTML]{F1F1F1} 0.338}} & \multicolumn{1}{l|}{\cellcolor[HTML]{070707}{\color[HTML]{F1F1F1} 0.343}} & \multicolumn{1}{l|}{\cellcolor[HTML]{343434}{\color[HTML]{F1F1F1} 0.438}} & \multicolumn{1}{l|}{\cellcolor[HTML]{252525}{\color[HTML]{F1F1F1} 0.406}} & \cellcolor[HTML]{242424}{\color[HTML]{F1F1F1} 0.405} \\ \cline{2-2} \cline{4-11} 
\multirow{-6}{*}{\textbf{30   minutes}} & \textbf{Transformer} & \multicolumn{1}{l|}{\multirow{-6}{*}{{\color[HTML]{F1F1F1} }}} & \multicolumn{1}{l|}{\cellcolor[HTML]{454545}{\color[HTML]{F1F1F1} 0.473}} & \multicolumn{1}{l|}{\cellcolor[HTML]{555555}{\color[HTML]{F1F1F1} 0.505}} & \multicolumn{1}{l|}{\cellcolor[HTML]{4D4D4D}{\color[HTML]{F1F1F1} 0.489}} & \multicolumn{1}{l|}{\cellcolor[HTML]{464646}{\color[HTML]{F1F1F1} 0.474}} & \multicolumn{1}{l|}{\cellcolor[HTML]{414141}{\color[HTML]{F1F1F1} 0.464}} & \multicolumn{1}{l|}{\cellcolor[HTML]{373737}{\color[HTML]{F1F1F1} 0.443}} & \multicolumn{1}{l|}{\cellcolor[HTML]{464646}{\color[HTML]{F1F1F1} 0.475}} & \cellcolor[HTML]{373737}{\color[HTML]{F1F1F1} 0.443} \\ \cline{1-2} \cline{4-11} 
 & \textbf{ConvLSTM} & {\color[HTML]{F1F1F1} } & \multicolumn{1}{l|}{{\color[HTML]{F1F1F1} }} & \multicolumn{1}{l|}{\cellcolor[HTML]{202020}{\color[HTML]{F1F1F1} 0.396}} & \multicolumn{1}{l|}{\cellcolor[HTML]{242424}{\color[HTML]{F1F1F1} 0.405}} & \multicolumn{1}{l|}{\cellcolor[HTML]{262626}{\color[HTML]{F1F1F1} 0.409}} & \multicolumn{1}{l|}{\cellcolor[HTML]{2B2B2B}{\color[HTML]{F1F1F1} 0.419}} & \multicolumn{1}{l|}{\cellcolor[HTML]{3F3F3F}{\color[HTML]{F1F1F1} 0.461}} & \multicolumn{1}{l|}{\cellcolor[HTML]{818181}{\color[HTML]{F1F1F1} 0.597}} & \cellcolor[HTML]{4E4E4E}{\color[HTML]{F1F1F1} 0.492} \\ \cline{2-2} \cline{5-11} 
 & \textbf{LR} & {\color[HTML]{F1F1F1} } & \multicolumn{1}{l|}{{\color[HTML]{F1F1F1} }} & \multicolumn{1}{l|}{\cellcolor[HTML]{272727}{\color[HTML]{F1F1F1} 0.410}} & \multicolumn{1}{l|}{\cellcolor[HTML]{1F1F1F}{\color[HTML]{F1F1F1} 0.393}} & \multicolumn{1}{l|}{\cellcolor[HTML]{1F1F1F}{\color[HTML]{F1F1F1} 0.393}} & \multicolumn{1}{l|}{\cellcolor[HTML]{202020}{\color[HTML]{F1F1F1} 0.395}} & \multicolumn{1}{l|}{\cellcolor[HTML]{454545}{\color[HTML]{F1F1F1} 0.473}} & \multicolumn{1}{l|}{\cellcolor[HTML]{5F5F5F}{\color[HTML]{F1F1F1} 0.526}} & \cellcolor[HTML]{4E4E4E}{\color[HTML]{F1F1F1} 0.492} \\ \cline{2-2} \cline{5-11} 
 & \textbf{LSTM} & {\color[HTML]{F1F1F1} } & \multicolumn{1}{l|}{{\color[HTML]{F1F1F1} }} & \multicolumn{1}{l|}{\cellcolor[HTML]{696969}{\color[HTML]{F1F1F1} 0.547}} & \multicolumn{1}{l|}{\cellcolor[HTML]{707070}{\color[HTML]{F1F1F1} 0.563}} & \multicolumn{1}{l|}{\cellcolor[HTML]{6E6E6E}{\color[HTML]{F1F1F1} 0.558}} & \multicolumn{1}{l|}{\cellcolor[HTML]{616161}{\color[HTML]{F1F1F1} 0.531}} & \multicolumn{1}{l|}{\cellcolor[HTML]{646464}{\color[HTML]{F1F1F1} 0.538}} & \multicolumn{1}{l|}{\cellcolor[HTML]{9A9A9A}{\color[HTML]{F1F1F1} 0.649}} & \cellcolor[HTML]{6E6E6E}{\color[HTML]{F1F1F1} 0.558} \\ \cline{2-2} \cline{5-11} 
 & \textbf{RF} & {\color[HTML]{F1F1F1} } & \multicolumn{1}{l|}{{\color[HTML]{F1F1F1} }} & \multicolumn{1}{l|}{\cellcolor[HTML]{020202}{\color[HTML]{F1F1F1} 0.333}} & \multicolumn{1}{l|}{\cellcolor[HTML]{020202}{\color[HTML]{F1F1F1} 0.333}} & \multicolumn{1}{l|}{\cellcolor[HTML]{020202}{\color[HTML]{F1F1F1} 0.333}} & \multicolumn{1}{l|}{\cellcolor[HTML]{020202}{\color[HTML]{F1F1F1} 0.333}} & \multicolumn{1}{l|}{\cellcolor[HTML]{010101}{\color[HTML]{F1F1F1} 0.332}} & \multicolumn{1}{l|}{\cellcolor[HTML]{0B0B0B}{\color[HTML]{F1F1F1} 0.351}} & \cellcolor[HTML]{070707}{\color[HTML]{F1F1F1} 0.343} \\ \cline{2-2} \cline{5-11} 
 & \textbf{SVM} & {\color[HTML]{F1F1F1} } & \multicolumn{1}{l|}{{\color[HTML]{F1F1F1} }} & \multicolumn{1}{l|}{\cellcolor[HTML]{010101}{\color[HTML]{F1F1F1} 0.328}} & \multicolumn{1}{l|}{\cellcolor[HTML]{010101}{\color[HTML]{F1F1F1} 0.328}} & \multicolumn{1}{l|}{\cellcolor[HTML]{010101}{\color[HTML]{F1F1F1} 0.328}} & \multicolumn{1}{l|}{\cellcolor[HTML]{010101}{\color[HTML]{F1F1F1} 0.329}} & \multicolumn{1}{l|}{\cellcolor[HTML]{161616}{\color[HTML]{F1F1F1} 0.375}} & \multicolumn{1}{l|}{\cellcolor[HTML]{303030}{\color[HTML]{F1F1F1} 0.428}} & \cellcolor[HTML]{040404}{\color[HTML]{F1F1F1} 0.338} \\ \cline{2-2} \cline{5-11} 
\multirow{-6}{*}{\textbf{35   minutes}} & \textbf{Transformer} & \multirow{-6}{*}{{\color[HTML]{F1F1F1} }} & \multicolumn{1}{l|}{\multirow{-6}{*}{{\color[HTML]{F1F1F1} }}} & \multicolumn{1}{l|}{\cellcolor[HTML]{5A5A5A}{\color[HTML]{F1F1F1} 0.516}} & \multicolumn{1}{l|}{\cellcolor[HTML]{444444}{\color[HTML]{F1F1F1} 0.471}} & \multicolumn{1}{l|}{\cellcolor[HTML]{4D4D4D}{\color[HTML]{F1F1F1} 0.490}} & \multicolumn{1}{l|}{\cellcolor[HTML]{414141}{\color[HTML]{F1F1F1} 0.464}} & \multicolumn{1}{l|}{\cellcolor[HTML]{323232}{\color[HTML]{F1F1F1} 0.433}} & \multicolumn{1}{l|}{\cellcolor[HTML]{4E4E4E}{\color[HTML]{F1F1F1} 0.492}} & \cellcolor[HTML]{414141}{\color[HTML]{F1F1F1} 0.464} \\ \hline
\end{tabular}%
}
\end{table}

\begin{figure}
    \centering
    \includegraphics[width=\textwidth]{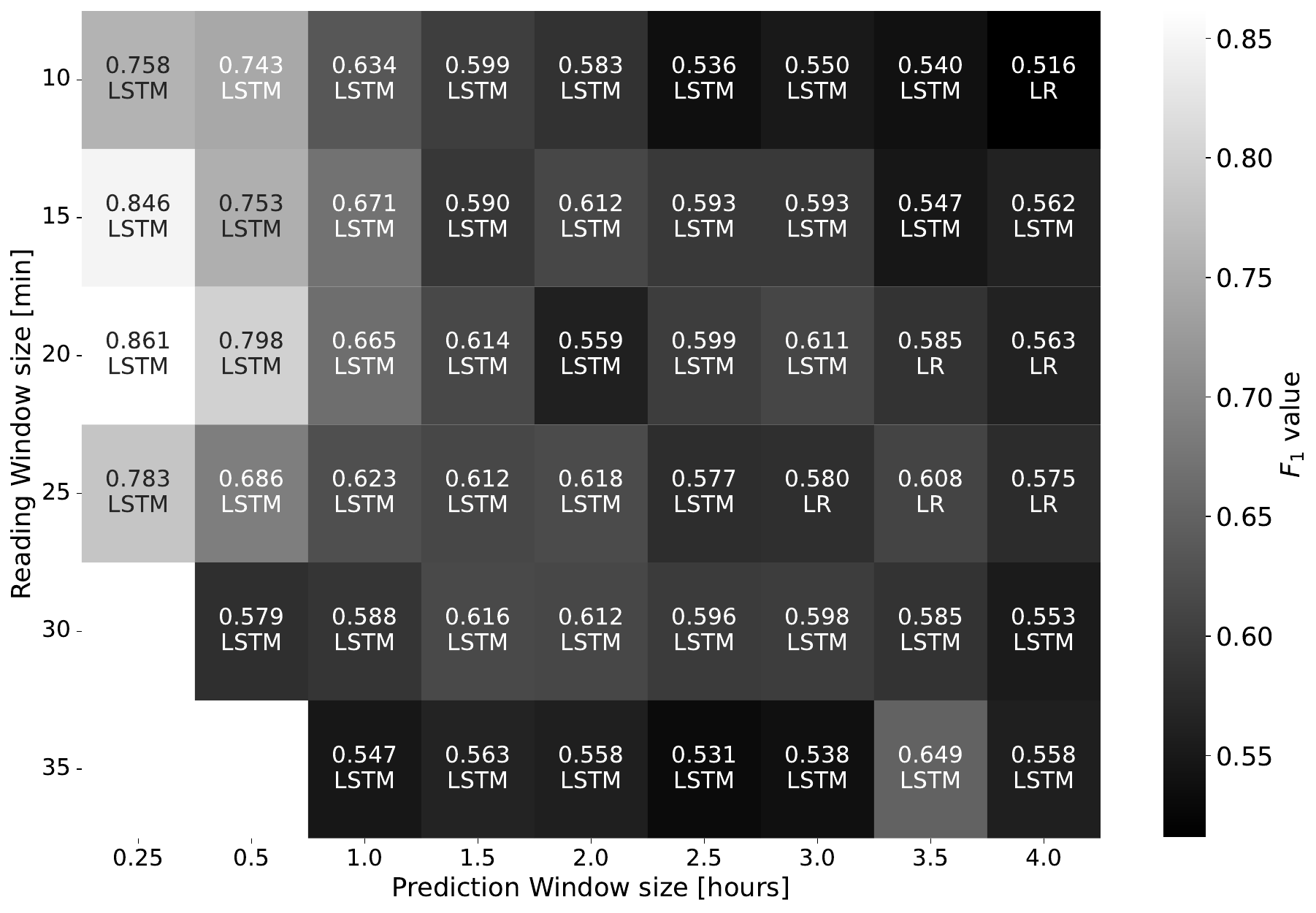}
    \caption{Comparison of the best $F_1$ scores for each combination of RW and PW sizes for the wrapping machine case study. The heatmap shows each combination's best method and the corresponding $F_1$ score. Lighter background colors correspond to better results.}
    \label{fig:max_f1_pw_rw}
\end{figure}

\textcolor{red}{In the wrapping machine case, the cardinality of the minority class samples is significantly influenced by both the PW and the RW, as reading windows spanning across multiple sessions are neglected.} Figure \ref{fig:support} shows the number of RWs with class "Failure" (i.e., the support) as RW and PW sizes vary. Note that a longer PW is more likely to contain a failure, resulting in a higher probability of the RW being classified as "Failure". Consequently, the number of RWs classified as "Failure" increases for larger PWs (i.e., the support increases). The support also increases with the decrease in RW size because more RWs are paired to a PW of class "Failure". 

Table \ref{tab:bc_f1_global_results} compares the $F_1$ scores of each algorithm as PW and RW vary (i.e., the $b_j$ values), and Figure \ref{fig:max_f1_pw_rw} shows the best algorithms for each reading and prediction window size (i.e., the $B_a$ values) and their $F_1$ scores (i.e., the $B_s$ values). In both cases, lighter backgrounds indicate better performances. Missing results correspond to cases in which at least one fold does not contain RWs of class "Failure".

\begin{figure}
    \centering
    \includegraphics[width=0.8\textwidth]{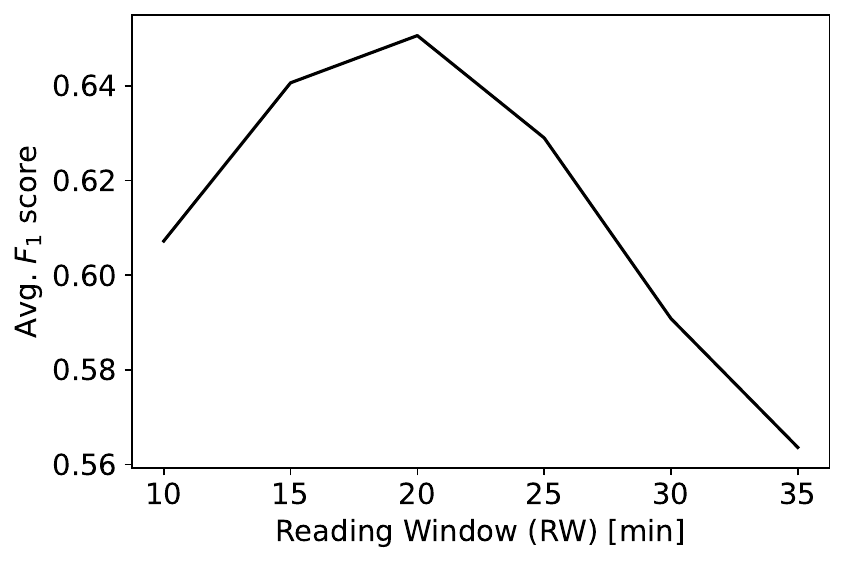}
    \caption{The $F_1$ score varies depending on the RW and, on average, reaches its peak for an RW of 20 minutes. This pattern is particularly noticeable for a PW of 0.25 hours.}
    \label{fig:reading_peak}
\end{figure}

Table \ref{tab:bc_f1_global_results}, Figure \ref{fig:max_f1_pw_rw} and Figure \ref{fig:reading_peak} show that the $F_1$ score, on average, increases as the RW size increases until a peak at 20 minutes and decreases afterward. Smaller train sets and the lesser relevance of past information can explain this behavior \cite{Zargoush2021, Nguyen2021}. For example, considering a PW of 15 minutes, the support increases of a factor $\approx 4$ from an RW of 30 minutes to an RW of 10 minutes. The work in  \cite{Nguyen2021} reports that decreasing the number of train set samples (in this case, the number of RWs in the train set) can reduce performances and that the best results are achieved for the greatest number of training samples. The work in \cite{Zargoush2021} finds that increasing the amount of historical data (in this case, the RW length) does not necessarily lead to better performance, as irrelevant information may be provided to the predictor. Similarly to \cite{Leukel2022}, the performances tend to decrease as the PW size increases, with the best performances observed for the smallest PW size. This pattern depends on the fact that the predictions for smaller PWs exploit more recent data than for larger PWs.

\begin{figure}[htb]
    \centering
    \includegraphics[width=0.7\textwidth]{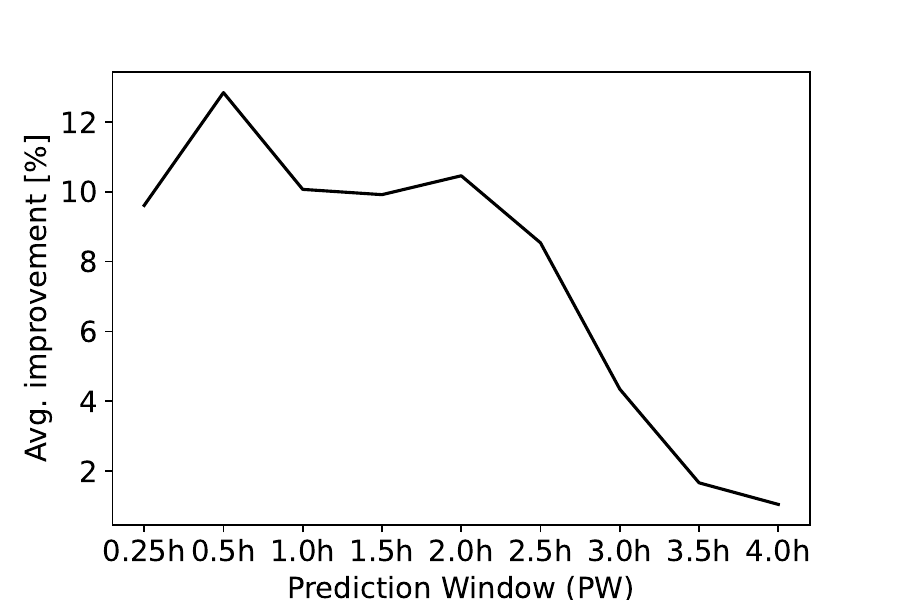}
    \caption{The average improvement of LSTM with respect to LR shows that LSTM benefits more from short PWs since it can capture relevant time-dependent patterns. Note that the values for the two smallest PWs (0.25 and 0.5 hours) are not entirely comparable with the others because they do not consider the longest RWs, for which the prediction could not be computed.}
    \label{fig:LSTM_improvement}
\end{figure}

LSTM performs better than the other algorithms (on average, $61.4\%$ $F_1$). \textcolor{red}{ConvLSTM is, on average, the second-best algorithm (on average, $54.5\%$ $F_1$), followed by LR (on average, $54.0\%$ $F_1$), Transformers (on average, $52.9\%$ $F_1$), SVM (on average, $47.4\%$ $F_1$), and RF (on average, $39.8\%$ $F_1$)}. Related studies on time series forecasting also show the effectiveness of LSTM over simpler non-DL approaches \cite{SiamiNamini2018, SiamiNamini2019, Rahimzad2021}. Figure \ref{fig:LSTM_improvement} shows that the improvement in performance introduced by LSTM \textcolor{red}{(the best DL algorithm)} with respect to LR \textcolor{red}{(the best ML algorithm)} is, on average, higher for smaller PWs, and lower for larger PWs. The work in \cite{Malakar2021}  also observes that LSTM forecasting performances decrease for a longer prediction horizon. This pattern can be explained by the decrease in the relevance of past information for predicting behaviors far in the future. In addition, \cite{Abduljabbar2021} observes that the performance improvement between different models tends to decrease as the forecasting horizon enlarges, suggesting that more complex approaches are as effective as less complex ones when the historical data are far in the past.  Overall, our findings show that in the case study, LSTM can capture temporal dependencies effectively and behave better for small PWs (i.e., the observed temporal patterns have a noticeable influence on the predicted class). As observed in \cite{Allam2019}, LR can exploit time-independent patterns since it does not consider any temporal dependency. This explains why the difference between LSTM and LR performances decreases as the PW increases.

Considering LSTM-based networks, the choice of the loss function also has an impact. As the RW and PW vary, the difficulty of predicting failures changes. In particular, predicting errors becomes harder when the PW increases, especially when less historical information is available (i.e., for small RWs and large PWs). In such cases, the $F_1$ loss is ineffective, as it maximizes the $F_1$ score for the "Failure" class. However, a relatively high score ($\approx 66.7\%$) corresponds to the trivial case in which only the "Failure" class is predicted, and it becomes more challenging to surpass this result in the most difficult instances of the prediction problem. Instead, the BCE loss does not privilege one of the two classes and is more suitable for the more challenging cases, leading to better macro $F_1$ scores.

\begin{table}[]
\renewcommand{\arraystretch}{1.3}
\centering
\caption{Comparison of the $F_1$ and BCE loss functions for LSTM. $F_1$ loss becomes less effective as the classification problem becomes more difficult (i.e. when the PW increases and the RW decreases)}
\label{tab:best-loss}
\resizebox{\textwidth}{!}{%
\begin{tabular}{|c|ccccccccc|}
\hline
 & \multicolumn{9}{c|}{\textbf{Prediction   Window (PW)}} \\ \cline{2-10} 
\multirow{-2}{*}{\textbf{Reading   Window (RW)}} & \multicolumn{1}{c|}{\textbf{0.25h}} & \multicolumn{1}{c|}{\textbf{0.5h}} & \multicolumn{1}{c|}{\textbf{1.0h}} & \multicolumn{1}{c|}{\textbf{1.5h}} & \multicolumn{1}{c|}{\textbf{2.0h}} & \multicolumn{1}{c|}{\textbf{2.5h}} & \multicolumn{1}{c|}{\textbf{3.0h}} & \multicolumn{1}{c|}{\textbf{3.5h}} & \textbf{4.0h} \\ \hline
\textbf{10 minutes} & \multicolumn{1}{c|}{$F_1$} & \multicolumn{1}{c|}{$F_1$} & \multicolumn{1}{c|}{\cellcolor[HTML]{D9D9D9}BCE} & \multicolumn{1}{c|}{\cellcolor[HTML]{D9D9D9}BCE} & \multicolumn{1}{c|}{\cellcolor[HTML]{D9D9D9}BCE} & \multicolumn{1}{c|}{\cellcolor[HTML]{D9D9D9}BCE} & \multicolumn{1}{c|}{\cellcolor[HTML]{D9D9D9}BCE} & \multicolumn{1}{c|}{\cellcolor[HTML]{D9D9D9}BCE} & \cellcolor[HTML]{D9D9D9}BCE \\ \hline
\textbf{15 minutes} & \multicolumn{1}{c|}{$F_1$} & \multicolumn{1}{c|}{$F_1$} & \multicolumn{1}{c|}{$F_1$} & \multicolumn{1}{c|}{\cellcolor[HTML]{D9D9D9}BCE} & \multicolumn{1}{c|}{\cellcolor[HTML]{D9D9D9}BCE} & \multicolumn{1}{c|}{\cellcolor[HTML]{D9D9D9}BCE} & \multicolumn{1}{c|}{\cellcolor[HTML]{D9D9D9}BCE} & \multicolumn{1}{c|}{\cellcolor[HTML]{D9D9D9}BCE} & \cellcolor[HTML]{D9D9D9}BCE \\ \hline
\textbf{20 minutes} & \multicolumn{1}{c|}{$F_1$} & \multicolumn{1}{c|}{$F_1$} & \multicolumn{1}{c|}{$F_1$} & \multicolumn{1}{c|}{$F_1$} & \multicolumn{1}{c|}{\cellcolor[HTML]{D9D9D9}BCE} & \multicolumn{1}{c|}{\cellcolor[HTML]{D9D9D9}BCE} & \multicolumn{1}{c|}{\cellcolor[HTML]{D9D9D9}BCE} & \multicolumn{1}{c|}{\cellcolor[HTML]{D9D9D9}BCE} & \cellcolor[HTML]{D9D9D9}BCE \\ \hline
\textbf{25 minutes} & \multicolumn{1}{c|}{$F_1$} & \multicolumn{1}{c|}{$F_1$} & \multicolumn{1}{c|}{$F_1$} & \multicolumn{1}{c|}{$F_1$} & \multicolumn{1}{c|}{$F_1$} & \multicolumn{1}{c|}{$F_1$} & \multicolumn{1}{c|}{\cellcolor[HTML]{D9D9D9}BCE} & \multicolumn{1}{c|}{\cellcolor[HTML]{D9D9D9}BCE} & \cellcolor[HTML]{D9D9D9}BCE \\ \hline
\textbf{30 minutes} & \multicolumn{1}{c|}{} & \multicolumn{1}{c|}{$F_1$} & \multicolumn{1}{c|}{$F_1$} & \multicolumn{1}{c|}{$F_1$} & \multicolumn{1}{c|}{$F_1$} & \multicolumn{1}{c|}{$F_1$} & \multicolumn{1}{c|}{$F_1$} & \multicolumn{1}{c|}{$F_1$} & $F_1$ \\ \hline
\textbf{35 minutes} & \multicolumn{1}{c|}{} & \multicolumn{1}{c|}{} & \multicolumn{1}{c|}{$F_1$} & \multicolumn{1}{c|}{$F_1$} & \multicolumn{1}{c|}{$F_1$} & \multicolumn{1}{c|}{$F_1$} & \multicolumn{1}{c|}{$F_1$} & \multicolumn{1}{c|}{$F_1$} & $F_1$ \\ \hline
\end{tabular}%
}
\end{table}

Focusing on the ML approaches, RF and SVM show poorer performances. SVM, similarly to LR, is not able to capture time dependencies. However, LR is expected to perform better than SVM, especially when the number of features exceeds the number of samples, as SVM with RBF is prone to overfitting \cite{Han2007}, while LR is a simpler model. In this case, the number of features for both algorithms is given by the number of samples in the reading windows multiplied by the number of time series variables. On average, there are more features than training windows ($\approx 1.13$ features per training sample), which justifies the better performances of LR over SVM ($\approx 1.14$ times better in terms of $F_1$ score, on average).

RF is the worst approach because each decision tree considers a limited number of features selected randomly, neglecting temporal dependencies, and the number of trees is much smaller than the number of input features ($\approx 2\%$ to $\approx 13\%$). For these reasons, the random forest ensembles weak estimators that cannot capture complex patterns. Increasing the number of estimators or the maximum number of features does not necessarily yield better results. For the combination of PW and RW leading to the best result for RF (i.e., 0.25 hours and 25 minutes, respectively), the best number of trees is 150, and the best number of maximum features is 33\% of all the features. Increasing the number of trees to 200 yields a -0.01\% variation in the macro $F_1$ score, and increasing the maximum number of features to 100\% yields a -0.1\% variation in the macro $F_1$ score. \textcolor{red}{On average across all the RW-PW combinations, the best ML algorithm has an $F_1$ of $\approx 54.8\%$, while the best DL algorithm (LSTM) has an $F_1$ of $\approx 61.4\%$. In this data set, DL algorithms result the best choice for the specified fault prediction task.}

In summary, the best results are obtained using a PW of 0.25 hours and an RW of 20 minutes, which is considered suitable in the industrial setting of the case study. A moderate amount of historical data is sufficient to predict an alert enough in advance to allow the operator to intervene. \textcolor{red}{For these RW and PW values, SVM, LR, LSTM, and ConvLSTM reach their highest $F_1$ scores (respectively 0.783, 0.789, 0.861, and 0.807)}. However, LSTM obtains the best $F_1$ score, which is $\approx 7\%$ better than LR and SVM, \textcolor{red}{$\approx 5\%$ better than ConvLSTM,} and $\approx 32\%$ better than RF. The difference between LSTM and the other approaches is justified by considering that \textcolor{red}{it can} consider temporal dependencies explicitly, and benefits are especially significant when recent historical data are used for predicting. \textcolor{red}{Regarding the other DL algorithms, Transformers focus on non-sequential patterns and  ConvLSTM on   spatial patterns, which makes them less effective for the specified time series fault prediction task.}

\subsection{Blood refrigerator}

\begin{table}[]
\centering
\caption{Classification results on the $F_1$ metrics for the blood refrigerator case study, varying both the reading and prediction window, for SVM, RF, LR, LSTM, Transformer, and ConvLSTM. Lighter background corresponds to better results.}
\renewcommand{\arraystretch}{1.1}
\label{tab:detail_blood_f1}
\begin{tabular}{|c|c|cccc|}
\hline
 &  & \multicolumn{4}{c|}{\textbf{Prediction   Window (PW)}} \\ \cline{3-6} 
\multirow{-2}{*}{\textbf{Reading   Window (RW)}} & \multirow{-2}{*}{\textbf{Algorithm}} & \multicolumn{1}{c|}{\textbf{0.5h}} & \multicolumn{1}{c|}{\textbf{1.0h}} & \multicolumn{1}{c|}{\textbf{1.5h}} & \textbf{2.0h} \\ \hline
 & \textbf{ConvLSTM} & \multicolumn{1}{c|}{\cellcolor[HTML]{C3C3C3}0.750} & \multicolumn{1}{c|}{\cellcolor[HTML]{606060}{\color[HTML]{F1F1F1} 0.612}} & \multicolumn{1}{c|}{\cellcolor[HTML]{707070}{\color[HTML]{F1F1F1} 0.634}} & \cellcolor[HTML]{464646}{\color[HTML]{F1F1F1} 0.575} \\ \cline{2-6} 
 & \textbf{LR} & \multicolumn{1}{c|}{\cellcolor[HTML]{B4B4B4}0.729} & \multicolumn{1}{c|}{\cellcolor[HTML]{595959}{\color[HTML]{F1F1F1} 0.602}} & \multicolumn{1}{c|}{\cellcolor[HTML]{5F5F5F}{\color[HTML]{F1F1F1} 0.610}} & \cellcolor[HTML]{595959}{\color[HTML]{F1F1F1} 0.602} \\ \cline{2-6} 
 & \textbf{LSTM} & \multicolumn{1}{c|}{\cellcolor[HTML]{B8B8B8}0.735} & \multicolumn{1}{c|}{\cellcolor[HTML]{626262}{\color[HTML]{F1F1F1} 0.615}} & \multicolumn{1}{c|}{\cellcolor[HTML]{6F6F6F}{\color[HTML]{F1F1F1} 0.633}} & \cellcolor[HTML]{6F6F6F}{\color[HTML]{F1F1F1} 0.633} \\ \cline{2-6} 
 & \textbf{RF} & \multicolumn{1}{c|}{\cellcolor[HTML]{CDCDCD}0.765} & \multicolumn{1}{c|}{\cellcolor[HTML]{414141}{\color[HTML]{F1F1F1} 0.568}} & \multicolumn{1}{c|}{\cellcolor[HTML]{212121}{\color[HTML]{F1F1F1} 0.524}} & \cellcolor[HTML]{313131}{\color[HTML]{F1F1F1} 0.546} \\ \cline{2-6} 
 & \textbf{SVM} & \multicolumn{1}{c|}{\cellcolor[HTML]{C1C1C1}0.748} & \multicolumn{1}{c|}{\cellcolor[HTML]{707070}{\color[HTML]{F1F1F1} 0.635}} & \multicolumn{1}{c|}{\cellcolor[HTML]{707070}{\color[HTML]{F1F1F1} 0.634}} & \cellcolor[HTML]{606060}{\color[HTML]{F1F1F1} 0.612} \\ \cline{2-6} 
\multirow{-6}{*}{\textbf{10   minutes}} & \textbf{Transformer} & \multicolumn{1}{c|}{\cellcolor[HTML]{E2E2E2}0.794} & \multicolumn{1}{c|}{\cellcolor[HTML]{666666}{\color[HTML]{F1F1F1} 0.621}} & \multicolumn{1}{c|}{\cellcolor[HTML]{393939}{\color[HTML]{F1F1F1} 0.558}} & \cellcolor[HTML]{828282}{\color[HTML]{F1F1F1} 0.659} \\ \hline
 & \textbf{ConvLSTM} & \multicolumn{1}{c|}{\cellcolor[HTML]{B2B2B2}0.727} & \multicolumn{1}{c|}{\cellcolor[HTML]{626262}{\color[HTML]{F1F1F1} 0.615}} & \multicolumn{1}{c|}{\cellcolor[HTML]{6F6F6F}{\color[HTML]{F1F1F1} 0.633}} & \cellcolor[HTML]{626262}{\color[HTML]{F1F1F1} 0.615} \\ \cline{2-6} 
 & \textbf{LR} & \multicolumn{1}{c|}{\cellcolor[HTML]{B2B2B2}0.727} & \multicolumn{1}{c|}{\cellcolor[HTML]{575757}{\color[HTML]{F1F1F1} 0.599}} & \multicolumn{1}{c|}{\cellcolor[HTML]{5D5D5D}{\color[HTML]{F1F1F1} 0.608}} & \cellcolor[HTML]{5D5D5D}{\color[HTML]{F1F1F1} 0.608} \\ \cline{2-6} 
 & \textbf{LSTM} & \multicolumn{1}{c|}{\cellcolor[HTML]{DADADA}0.782} & \multicolumn{1}{c|}{\cellcolor[HTML]{808080}{\color[HTML]{F1F1F1} 0.657}} & \multicolumn{1}{c|}{\cellcolor[HTML]{575757}{\color[HTML]{F1F1F1} 0.599}} & \cellcolor[HTML]{3D3D3D}{\color[HTML]{F1F1F1} 0.563} \\ \cline{2-6} 
 & \textbf{RF} & \multicolumn{1}{c|}{\cellcolor[HTML]{D3D3D3}0.773} & \multicolumn{1}{c|}{\cellcolor[HTML]{696969}{\color[HTML]{F1F1F1} 0.624}} & \multicolumn{1}{c|}{\cellcolor[HTML]{414141}{\color[HTML]{F1F1F1} 0.569}} & \cellcolor[HTML]{323232}{\color[HTML]{F1F1F1} 0.548} \\ \cline{2-6} 
 & \textbf{SVM} & \multicolumn{1}{c|}{\cellcolor[HTML]{BDBDBD}0.742} & \multicolumn{1}{c|}{\cellcolor[HTML]{7B7B7B}{\color[HTML]{F1F1F1} 0.650}} & \multicolumn{1}{c|}{\cellcolor[HTML]{747474}{\color[HTML]{F1F1F1} 0.640}} & \cellcolor[HTML]{5E5E5E}{\color[HTML]{F1F1F1} 0.609} \\ \cline{2-6} 
\multirow{-6}{*}{\textbf{15   minutes}} & \textbf{Transformer} & \multicolumn{1}{c|}{\cellcolor[HTML]{DDDDDD}0.787} & \multicolumn{1}{c|}{\cellcolor[HTML]{343434}{\color[HTML]{F1F1F1} 0.551}} & \multicolumn{1}{c|}{\cellcolor[HTML]{383838}{\color[HTML]{F1F1F1} 0.556}} & \cellcolor[HTML]{5A5A5A}{\color[HTML]{F1F1F1} 0.603} \\ \hline
 & \textbf{ConvLSTM} & \multicolumn{1}{c|}{\cellcolor[HTML]{919191}{\color[HTML]{F1F1F1} 0.680}} & \multicolumn{1}{c|}{\cellcolor[HTML]{696969}{\color[HTML]{F1F1F1} 0.624}} & \multicolumn{1}{c|}{\cellcolor[HTML]{646464}{\color[HTML]{F1F1F1} 0.617}} & \cellcolor[HTML]{464646}{\color[HTML]{F1F1F1} 0.576} \\ \cline{2-6} 
 & \textbf{LR} & \multicolumn{1}{c|}{\cellcolor[HTML]{A8A8A8}{\color[HTML]{F1F1F1} 0.712}} & \multicolumn{1}{c|}{\cellcolor[HTML]{5B5B5B}{\color[HTML]{F1F1F1} 0.605}} & \multicolumn{1}{c|}{\cellcolor[HTML]{565656}{\color[HTML]{F1F1F1} 0.598}} & \cellcolor[HTML]{5D5D5D}{\color[HTML]{F1F1F1} 0.608} \\ \cline{2-6} 
 & \textbf{LSTM} & \multicolumn{1}{c|}{\cellcolor[HTML]{CACACA}0.760} & \multicolumn{1}{c|}{\cellcolor[HTML]{767676}{\color[HTML]{F1F1F1} 0.643}} & \multicolumn{1}{c|}{\cellcolor[HTML]{5C5C5C}{\color[HTML]{F1F1F1} 0.607}} & \cellcolor[HTML]{414141}{\color[HTML]{F1F1F1} 0.568} \\ \cline{2-6} 
 & \textbf{RF} & \multicolumn{1}{c|}{\cellcolor[HTML]{F0F0F0}0.814} & \multicolumn{1}{c|}{\cellcolor[HTML]{656565}{\color[HTML]{F1F1F1} 0.619}} & \multicolumn{1}{c|}{\cellcolor[HTML]{454545}{\color[HTML]{F1F1F1} 0.574}} & \cellcolor[HTML]{343434}{\color[HTML]{F1F1F1} 0.550} \\ \cline{2-6} 
 & \textbf{SVM} & \multicolumn{1}{c|}{\cellcolor[HTML]{BEBEBE}0.744} & \multicolumn{1}{c|}{\cellcolor[HTML]{7A7A7A}{\color[HTML]{F1F1F1} 0.648}} & \multicolumn{1}{c|}{\cellcolor[HTML]{717171}{\color[HTML]{F1F1F1} 0.636}} & \cellcolor[HTML]{616161}{\color[HTML]{F1F1F1} 0.614} \\ \cline{2-6} 
\multirow{-6}{*}{\textbf{20   minutes}} & \textbf{Transformer} & \multicolumn{1}{c|}{\cellcolor[HTML]{E3E3E3}0.795} & \multicolumn{1}{c|}{\cellcolor[HTML]{6F6F6F}{\color[HTML]{F1F1F1} 0.633}} & \multicolumn{1}{c|}{\cellcolor[HTML]{8A8A8A}{\color[HTML]{F1F1F1} 0.670}} & \cellcolor[HTML]{828282}{\color[HTML]{F1F1F1} 0.659} \\ \hline
 & \textbf{ConvLSTM} & \multicolumn{1}{c|}{\cellcolor[HTML]{B1B1B1}0.725} & \multicolumn{1}{c|}{\cellcolor[HTML]{707070}{\color[HTML]{F1F1F1} 0.634}} & \multicolumn{1}{c|}{\cellcolor[HTML]{6B6B6B}{\color[HTML]{F1F1F1} 0.627}} & \cellcolor[HTML]{2E2E2E}{\color[HTML]{F1F1F1} 0.542} \\ \cline{2-6} 
 & \textbf{LR} & \multicolumn{1}{c|}{\cellcolor[HTML]{AEAEAE}0.721} & \multicolumn{1}{c|}{\cellcolor[HTML]{5B5B5B}{\color[HTML]{F1F1F1} 0.605}} & \multicolumn{1}{c|}{\cellcolor[HTML]{575757}{\color[HTML]{F1F1F1} 0.599}} & \cellcolor[HTML]{575757}{\color[HTML]{F1F1F1} 0.599} \\ \cline{2-6} 
 & \textbf{LSTM} & \multicolumn{1}{c|}{\cellcolor[HTML]{D7D7D7}0.779} & \multicolumn{1}{c|}{\cellcolor[HTML]{868686}{\color[HTML]{F1F1F1} 0.665}} & \multicolumn{1}{c|}{\cellcolor[HTML]{1E1E1E}{\color[HTML]{F1F1F1} 0.520}} & \cellcolor[HTML]{080808}{\color[HTML]{F1F1F1} 0.489} \\ \cline{2-6} 
 & \textbf{RF} & \multicolumn{1}{c|}{\cellcolor[HTML]{F6F6F6}0.822} & \multicolumn{1}{c|}{\cellcolor[HTML]{575757}{\color[HTML]{F1F1F1} 0.599}} & \multicolumn{1}{c|}{\cellcolor[HTML]{484848}{\color[HTML]{F1F1F1} 0.578}} & \cellcolor[HTML]{4D4D4D}{\color[HTML]{F1F1F1} 0.586} \\ \cline{2-6} 
 & \textbf{SVM} & \multicolumn{1}{c|}{\cellcolor[HTML]{BABABA}0.738} & \multicolumn{1}{c|}{\cellcolor[HTML]{7C7C7C}{\color[HTML]{F1F1F1} 0.651}} & \multicolumn{1}{c|}{\cellcolor[HTML]{707070}{\color[HTML]{F1F1F1} 0.634}} & \cellcolor[HTML]{616161}{\color[HTML]{F1F1F1} 0.613} \\ \cline{2-6} 
\multirow{-6}{*}{\textbf{25   minutes}} & \textbf{Transformer} & \multicolumn{1}{c|}{\cellcolor[HTML]{C8C8C8}0.758} & \multicolumn{1}{c|}{\cellcolor[HTML]{818181}{\color[HTML]{F1F1F1} 0.658}} & \multicolumn{1}{c|}{\cellcolor[HTML]{5C5C5C}{\color[HTML]{F1F1F1} 0.607}} & \cellcolor[HTML]{8A8A8A}{\color[HTML]{F1F1F1} 0.670} \\ \hline
 & \textbf{ConvLSTM} & \multicolumn{1}{c|}{\cellcolor[HTML]{B4B4B4}0.730} & \multicolumn{1}{c|}{\cellcolor[HTML]{2F2F2F}{\color[HTML]{F1F1F1} 0.544}} & \multicolumn{1}{c|}{\cellcolor[HTML]{585858}{\color[HTML]{F1F1F1} 0.601}} & \cellcolor[HTML]{4F4F4F}{\color[HTML]{F1F1F1} 0.588} \\ \cline{2-6} 
 & \textbf{LR} & \multicolumn{1}{c|}{\cellcolor[HTML]{B0B0B0}0.724} & \multicolumn{1}{c|}{\cellcolor[HTML]{565656}{\color[HTML]{F1F1F1} 0.598}} & \multicolumn{1}{c|}{\cellcolor[HTML]{5A5A5A}{\color[HTML]{F1F1F1} 0.603}} & \cellcolor[HTML]{4E4E4E}{\color[HTML]{F1F1F1} 0.587} \\ \cline{2-6} 
 & \textbf{LSTM} & \multicolumn{1}{c|}{\cellcolor[HTML]{E9E9E9}0.804} & \multicolumn{1}{c|}{\cellcolor[HTML]{6F6F6F}{\color[HTML]{F1F1F1} 0.633}} & \multicolumn{1}{c|}{\cellcolor[HTML]{474747}{\color[HTML]{F1F1F1} 0.577}} & \cellcolor[HTML]{010101}{\color[HTML]{F1F1F1} 0.477} \\ \cline{2-6} 
 & \textbf{RF} & \multicolumn{1}{c|}{\cellcolor[HTML]{FFFFFF}0.835} & \multicolumn{1}{c|}{\cellcolor[HTML]{5C5C5C}{\color[HTML]{F1F1F1} 0.606}} & \multicolumn{1}{c|}{\cellcolor[HTML]{434343}{\color[HTML]{F1F1F1} 0.571}} & \cellcolor[HTML]{373737}{\color[HTML]{F1F1F1} 0.555} \\ \cline{2-6} 
 & \textbf{SVM} & \multicolumn{1}{c|}{\cellcolor[HTML]{BDBDBD}0.742} & \multicolumn{1}{c|}{\cellcolor[HTML]{707070}{\color[HTML]{F1F1F1} 0.635}} & \multicolumn{1}{c|}{\cellcolor[HTML]{6E6E6E}{\color[HTML]{F1F1F1} 0.632}} & \cellcolor[HTML]{5C5C5C}{\color[HTML]{F1F1F1} 0.606} \\ \cline{2-6} 
\multirow{-6}{*}{\textbf{30   minutes}} & \textbf{Transformer} & \multicolumn{1}{c|}{\cellcolor[HTML]{B3B3B3}0.728} & \multicolumn{1}{c|}{\cellcolor[HTML]{787878}{\color[HTML]{F1F1F1} 0.646}} & \multicolumn{1}{c|}{\cellcolor[HTML]{616161}{\color[HTML]{F1F1F1} 0.613}} & \cellcolor[HTML]{1E1E1E}{\color[HTML]{F1F1F1} 0.519} \\ \hline
\end{tabular}
\end{table}

\textcolor{red}{Table \ref{tab:detail_blood_f1} shows the $F_1$ scores of each algorithm as PW and RW vary, similarly to Table \ref{tab:bc_f1_global_results}. DL algorithms outperform ML algorithms in less cases than in the wrapping machine time series. Such difference can be explained by considering the different nature of the data sets. While the wrapping machine faults are not associated with clearly identifiable short patterns and are caused by a longer wearing of the machine components, in the blood refrigerator, the high product temperature is often observed as a consequence of some short easily identifiable patterns, exemplified in Figure \ref{fig:blood_patterns}. Such patterns can be  found by simpler algorithms (i.e., ML algorithms), and more complex architectures do not lead to significant benefits.}

\begin{figure}
    \centering
    \includegraphics[width=\textwidth]{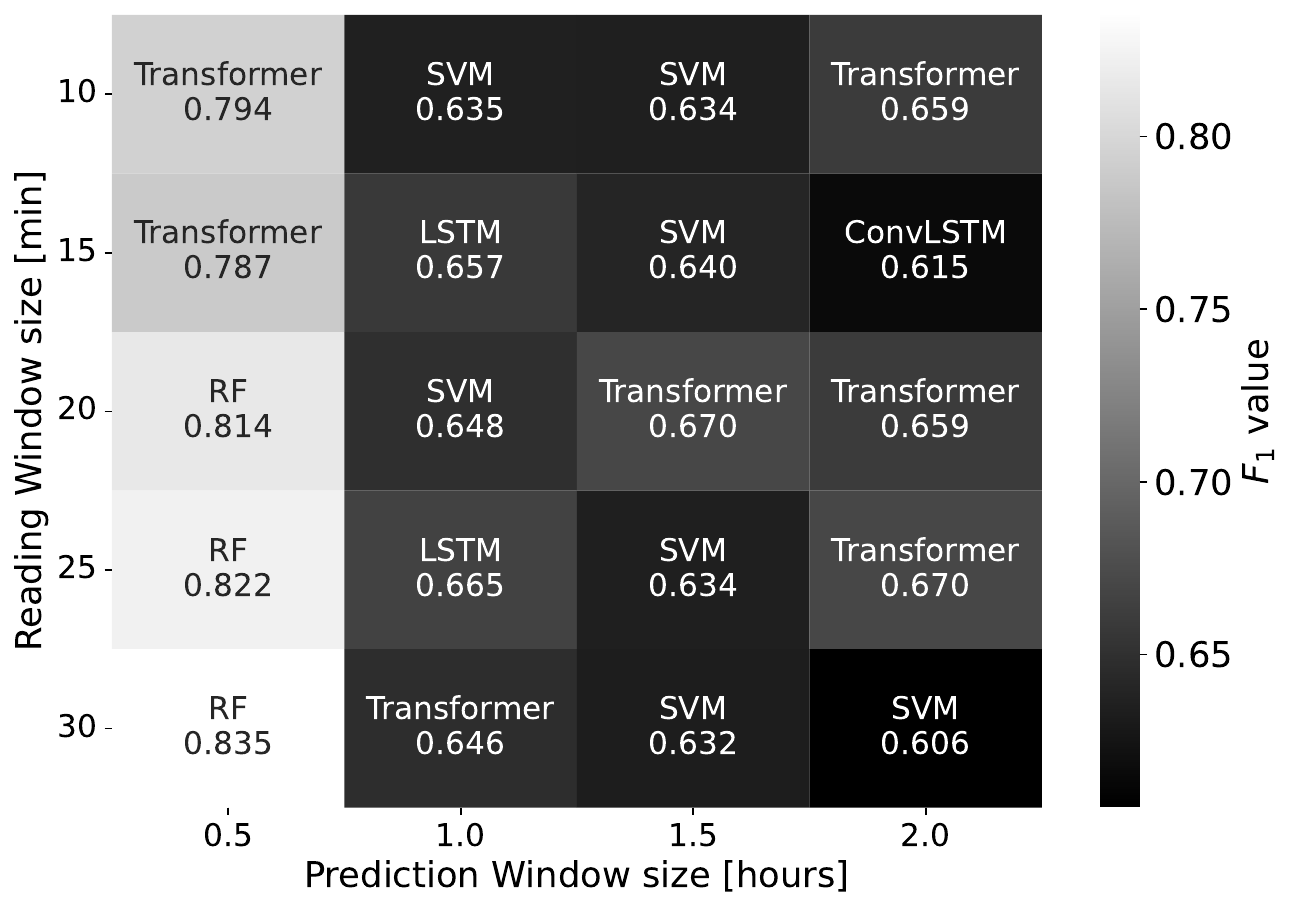}
    \caption{Comparison of the best $F_1$ scores for each combination of RW and PW sizes for the blood refrigerator case study. The heatmap shows each combination's best method and the corresponding $F_1$ score. Lighter background colors correspond to better results.}
    \label{fig:blood_f1_best}
\end{figure}

\textcolor{red}{
Figure \ref{fig:blood_f1_best} presents the algorithm with the best $F_1$ score for each RW and PW pair. The  results highlight the small variation of  performances across the  algorithms. On average, the best ML algorithm for each RW-PW combination has an $F_1$ of $\approx 67.3\%$, while the best DL algorithm for each RW-PW combination has an $F_1$ of $\approx 67.8\%$. }

\textcolor{red}{In the case of the blood refrigerator, the absence of sessions leads to negligible differences in the support, and does not affect results significantly. In this case, the average $F_1$ score for different RWs varies between $\approx 67.4\%$ and $\approx 69.8\%$, a small variation ($\approx 2.4\%$) compared to the one of the wrapping machine data set ($\approx 9\%$). This result also suggests that the amount of historical data has, for this data set, a negligible effect, and that short RWs already lead to high $F_1$ scores.}

\subsection{Nitrogen generator}

\begin{table}[]
\centering
\caption{Classification results on the $F_1$ metrics for the nitrogen generator case study, varying both the reading and prediction window, for SVM, RF, LR, LSTM, Transformer, and ConvLSTM. Lighter background corresponds to better results.}
\label{tab:all_f1_nitrogen}
\begin{tabular}{|c|c|cccccc|}
\hline
 &  & \multicolumn{6}{c|}{\textbf{Prediction   Window (PW)}} \\ \cline{3-8} 
\multirow{-2}{*}{\textbf{Reading   Window (RW)}} & \multirow{-2}{*}{\textbf{Algorithm}} & \multicolumn{1}{c|}{\textbf{0.5h}} & \multicolumn{1}{c|}{\textbf{1.0h}} & \multicolumn{1}{c|}{\textbf{1.5h}} & \multicolumn{1}{c|}{\textbf{2.0h}} & \multicolumn{1}{c|}{\textbf{3.0h}} & \textbf{5.0h} \\ \hline
 & \textbf{ConvLSTM} & \multicolumn{1}{c|}{\cellcolor[HTML]{CBCBCB}0.875} & \multicolumn{1}{c|}{\cellcolor[HTML]{E2E2E2}0.900} & \multicolumn{1}{c|}{\cellcolor[HTML]{C9C9C9}0.873} & \multicolumn{1}{c|}{\cellcolor[HTML]{CECECE}0.879} & \multicolumn{1}{c|}{\cellcolor[HTML]{8D8D8D}{\color[HTML]{F1F1F1} 0.810}} & \cellcolor[HTML]{7C7C7C}{\color[HTML]{F1F1F1} 0.792} \\ \cline{2-8} 
 & \textbf{LR} & \multicolumn{1}{c|}{\cellcolor[HTML]{8F8F8F}{\color[HTML]{F1F1F1} 0.812}} & \multicolumn{1}{c|}{\cellcolor[HTML]{9A9A9A}{\color[HTML]{F1F1F1} 0.824}} & \multicolumn{1}{c|}{\cellcolor[HTML]{AAAAAA}{\color[HTML]{F1F1F1} 0.841}} & \multicolumn{1}{c|}{\cellcolor[HTML]{A5A5A5}{\color[HTML]{F1F1F1} 0.835}} & \multicolumn{1}{c|}{\cellcolor[HTML]{010101}{\color[HTML]{F1F1F1} 0.660}} & \cellcolor[HTML]{0D0D0D}{\color[HTML]{F1F1F1} 0.674} \\ \cline{2-8} 
 & \textbf{LSTM} & \multicolumn{1}{c|}{\cellcolor[HTML]{CACACA}0.874} & \multicolumn{1}{c|}{\cellcolor[HTML]{C8C8C8}0.872} & \multicolumn{1}{c|}{\cellcolor[HTML]{C5C5C5}0.869} & \multicolumn{1}{c|}{\cellcolor[HTML]{B9B9B9}0.856} & \multicolumn{1}{c|}{\cellcolor[HTML]{C1C1C1}0.865} & \cellcolor[HTML]{898989}{\color[HTML]{F1F1F1} 0.806} \\ \cline{2-8} 
 & \textbf{RF} & \multicolumn{1}{c|}{\cellcolor[HTML]{FFFFFF}0.931} & \multicolumn{1}{c|}{\cellcolor[HTML]{D4D4D4}0.885} & \multicolumn{1}{c|}{\cellcolor[HTML]{D2D2D2}0.883} & \multicolumn{1}{c|}{\cellcolor[HTML]{D2D2D2}0.883} & \multicolumn{1}{c|}{\cellcolor[HTML]{ADADAD}0.844} & \cellcolor[HTML]{A8A8A8}{\color[HTML]{F1F1F1} 0.838} \\ \cline{2-8} 
 & \textbf{SVM} & \multicolumn{1}{c|}{\cellcolor[HTML]{E5E5E5}0.903} & \multicolumn{1}{c|}{\cellcolor[HTML]{D1D1D1}0.882} & \multicolumn{1}{c|}{\cellcolor[HTML]{D4D4D4}0.885} & \multicolumn{1}{c|}{\cellcolor[HTML]{C7C7C7}0.871} & \multicolumn{1}{c|}{\cellcolor[HTML]{8A8A8A}{\color[HTML]{F1F1F1} 0.807}} & \cellcolor[HTML]{686868}{\color[HTML]{F1F1F1} 0.771} \\ \cline{2-8} 
\multirow{-6}{*}{\textbf{10   minutes}} & \textbf{Transformer} & \multicolumn{1}{c|}{\cellcolor[HTML]{CACACA}0.874} & \multicolumn{1}{c|}{\cellcolor[HTML]{CECECE}0.879} & \multicolumn{1}{c|}{\cellcolor[HTML]{C7C7C7}0.871} & \multicolumn{1}{c|}{\cellcolor[HTML]{D9D9D9}0.890} & \multicolumn{1}{c|}{\cellcolor[HTML]{9E9E9E}{\color[HTML]{F1F1F1} 0.828}} & \cellcolor[HTML]{8E8E8E}{\color[HTML]{F1F1F1} 0.811} \\ \hline
 & \textbf{ConvLSTM} & \multicolumn{1}{c|}{\cellcolor[HTML]{A0A0A0}{\color[HTML]{F1F1F1} 0.830}} & \multicolumn{1}{c|}{\cellcolor[HTML]{E6E6E6}0.904} & \multicolumn{1}{c|}{\cellcolor[HTML]{828282}{\color[HTML]{F1F1F1} 0.798}} & \multicolumn{1}{c|}{\cellcolor[HTML]{C4C4C4}0.868} & \multicolumn{1}{c|}{\cellcolor[HTML]{A0A0A0}{\color[HTML]{F1F1F1} 0.830}} & \cellcolor[HTML]{B1B1B1}0.848 \\ \cline{2-8} 
 & \textbf{LR} & \multicolumn{1}{c|}{\cellcolor[HTML]{A7A7A7}{\color[HTML]{F1F1F1} 0.837}} & \multicolumn{1}{c|}{\cellcolor[HTML]{A7A7A7}{\color[HTML]{F1F1F1} 0.837}} & \multicolumn{1}{c|}{\cellcolor[HTML]{AAAAAA}{\color[HTML]{F1F1F1} 0.840}} & \multicolumn{1}{c|}{\cellcolor[HTML]{9E9E9E}{\color[HTML]{F1F1F1} 0.828}} & \multicolumn{1}{c|}{\cellcolor[HTML]{020202}{\color[HTML]{F1F1F1} 0.663}} & \cellcolor[HTML]{0F0F0F}{\color[HTML]{F1F1F1} 0.676} \\ \cline{2-8} 
 & \textbf{LSTM} & \multicolumn{1}{c|}{\cellcolor[HTML]{959595}{\color[HTML]{F1F1F1} 0.818}} & \multicolumn{1}{c|}{\cellcolor[HTML]{DBDBDB}0.892} & \multicolumn{1}{c|}{\cellcolor[HTML]{D6D6D6}0.887} & \multicolumn{1}{c|}{\cellcolor[HTML]{CECECE}0.879} & \multicolumn{1}{c|}{\cellcolor[HTML]{A0A0A0}{\color[HTML]{F1F1F1} 0.830}} & \cellcolor[HTML]{8A8A8A}{\color[HTML]{F1F1F1} 0.807} \\ \cline{2-8} 
 & \textbf{RF} & \multicolumn{1}{c|}{\cellcolor[HTML]{F4F4F4}0.919} & \multicolumn{1}{c|}{\cellcolor[HTML]{DBDBDB}0.892} & \multicolumn{1}{c|}{\cellcolor[HTML]{CCCCCC}0.876} & \multicolumn{1}{c|}{\cellcolor[HTML]{CDCDCD}0.878} & \multicolumn{1}{c|}{\cellcolor[HTML]{A4A4A4}{\color[HTML]{F1F1F1} 0.834}} & \cellcolor[HTML]{A8A8A8}{\color[HTML]{F1F1F1} 0.838} \\ \cline{2-8} 
 & \textbf{SVM} & \multicolumn{1}{c|}{\cellcolor[HTML]{DDDDDD}0.894} & \multicolumn{1}{c|}{\cellcolor[HTML]{CECECE}0.879} & \multicolumn{1}{c|}{\cellcolor[HTML]{CBCBCB}0.875} & \multicolumn{1}{c|}{\cellcolor[HTML]{CACACA}0.874} & \multicolumn{1}{c|}{\cellcolor[HTML]{8D8D8D}{\color[HTML]{F1F1F1} 0.810}} & \cellcolor[HTML]{757575}{\color[HTML]{F1F1F1} 0.784} \\ \cline{2-8} 
\multirow{-6}{*}{\textbf{15   minutes}} & \textbf{Transformer} & \multicolumn{1}{c|}{\cellcolor[HTML]{DBDBDB}0.892} & \multicolumn{1}{c|}{\cellcolor[HTML]{D7D7D7}0.888} & \multicolumn{1}{c|}{\cellcolor[HTML]{DCDCDC}0.893} & \multicolumn{1}{c|}{\cellcolor[HTML]{D2D2D2}0.883} & \multicolumn{1}{c|}{\cellcolor[HTML]{A9A9A9}{\color[HTML]{F1F1F1} 0.839}} & \cellcolor[HTML]{989898}{\color[HTML]{F1F1F1} 0.821} \\ \hline
 & \textbf{ConvLSTM} & \multicolumn{1}{c|}{\cellcolor[HTML]{D7D7D7}0.888} & \multicolumn{1}{c|}{\cellcolor[HTML]{DCDCDC}0.893} & \multicolumn{1}{c|}{\cellcolor[HTML]{D1D1D1}0.882} & \multicolumn{1}{c|}{\cellcolor[HTML]{CCCCCC}0.876} & \multicolumn{1}{c|}{\cellcolor[HTML]{B9B9B9}0.856} & \cellcolor[HTML]{8F8F8F}{\color[HTML]{F1F1F1} 0.812} \\ \cline{2-8} 
 & \textbf{LR} & \multicolumn{1}{c|}{\cellcolor[HTML]{A2A2A2}{\color[HTML]{F1F1F1} 0.832}} & \multicolumn{1}{c|}{\cellcolor[HTML]{A7A7A7}{\color[HTML]{F1F1F1} 0.837}} & \multicolumn{1}{c|}{\cellcolor[HTML]{AAAAAA}{\color[HTML]{F1F1F1} 0.841}} & \multicolumn{1}{c|}{\cellcolor[HTML]{9C9C9C}{\color[HTML]{F1F1F1} 0.826}} & \multicolumn{1}{c|}{\cellcolor[HTML]{090909}{\color[HTML]{F1F1F1} 0.670}} & \cellcolor[HTML]{111111}{\color[HTML]{F1F1F1} 0.678} \\ \cline{2-8} 
 & \textbf{LSTM} & \multicolumn{1}{c|}{\cellcolor[HTML]{A4A4A4}{\color[HTML]{F1F1F1} 0.834}} & \multicolumn{1}{c|}{\cellcolor[HTML]{D4D4D4}0.885} & \multicolumn{1}{c|}{\cellcolor[HTML]{E8E8E8}0.906} & \multicolumn{1}{c|}{\cellcolor[HTML]{CDCDCD}0.878} & \multicolumn{1}{c|}{\cellcolor[HTML]{909090}{\color[HTML]{F1F1F1} 0.813}} & \cellcolor[HTML]{919191}{\color[HTML]{F1F1F1} 0.814} \\ \cline{2-8} 
 & \textbf{RF} & \multicolumn{1}{c|}{\cellcolor[HTML]{E4E4E4}0.902} & \multicolumn{1}{c|}{\cellcolor[HTML]{D9D9D9}0.890} & \multicolumn{1}{c|}{\cellcolor[HTML]{C9C9C9}0.873} & \multicolumn{1}{c|}{\cellcolor[HTML]{D1D1D1}0.882} & \multicolumn{1}{c|}{\cellcolor[HTML]{ACACAC}0.843} & \cellcolor[HTML]{A4A4A4}{\color[HTML]{F1F1F1} 0.834} \\ \cline{2-8} 
 & \textbf{SVM} & \multicolumn{1}{c|}{\cellcolor[HTML]{D1D1D1}0.882} & \multicolumn{1}{c|}{\cellcolor[HTML]{CCCCCC}0.877} & \multicolumn{1}{c|}{\cellcolor[HTML]{C9C9C9}0.873} & \multicolumn{1}{c|}{\cellcolor[HTML]{D0D0D0}0.881} & \multicolumn{1}{c|}{\cellcolor[HTML]{929292}{\color[HTML]{F1F1F1} 0.815}} & \cellcolor[HTML]{7F7F7F}{\color[HTML]{F1F1F1} 0.795} \\ \cline{2-8} 
\multirow{-6}{*}{\textbf{20   minutes}} & \textbf{Transformer} & \multicolumn{1}{c|}{\cellcolor[HTML]{DEDEDE}0.896} & \multicolumn{1}{c|}{\cellcolor[HTML]{DADADA}0.891} & \multicolumn{1}{c|}{\cellcolor[HTML]{DEDEDE}0.896} & \multicolumn{1}{c|}{\cellcolor[HTML]{D7D7D7}0.888} & \multicolumn{1}{c|}{\cellcolor[HTML]{909090}{\color[HTML]{F1F1F1} 0.813}} & \cellcolor[HTML]{9B9B9B}{\color[HTML]{F1F1F1} 0.825} \\ \hline
 & \textbf{ConvLSTM} & \multicolumn{1}{c|}{\cellcolor[HTML]{ABABAB}{\color[HTML]{F1F1F1} 0.842}} & \multicolumn{1}{c|}{\cellcolor[HTML]{C8C8C8}0.872} & \multicolumn{1}{c|}{\cellcolor[HTML]{D6D6D6}0.887} & \multicolumn{1}{c|}{\cellcolor[HTML]{D9D9D9}0.890} & \multicolumn{1}{c|}{\cellcolor[HTML]{8C8C8C}{\color[HTML]{F1F1F1} 0.809}} & \cellcolor[HTML]{999999}{\color[HTML]{F1F1F1} 0.823} \\ \cline{2-8} 
 & \textbf{LR} & \multicolumn{1}{c|}{\cellcolor[HTML]{AEAEAE}0.845} & \multicolumn{1}{c|}{\cellcolor[HTML]{A8A8A8}{\color[HTML]{F1F1F1} 0.838}} & \multicolumn{1}{c|}{\cellcolor[HTML]{B7B7B7}0.854} & \multicolumn{1}{c|}{\cellcolor[HTML]{9D9D9D}{\color[HTML]{F1F1F1} 0.827}} & \multicolumn{1}{c|}{\cellcolor[HTML]{0E0E0E}{\color[HTML]{F1F1F1} 0.675}} & \cellcolor[HTML]{121212}{\color[HTML]{F1F1F1} 0.680} \\ \cline{2-8} 
 & \textbf{LSTM} & \multicolumn{1}{c|}{\cellcolor[HTML]{919191}{\color[HTML]{F1F1F1} 0.814}} & \multicolumn{1}{c|}{\cellcolor[HTML]{DCDCDC}0.893} & \multicolumn{1}{c|}{\cellcolor[HTML]{D7D7D7}0.888} & \multicolumn{1}{c|}{\cellcolor[HTML]{CECECE}0.879} & \multicolumn{1}{c|}{\cellcolor[HTML]{A2A2A2}{\color[HTML]{F1F1F1} 0.832}} & \cellcolor[HTML]{8E8E8E}{\color[HTML]{F1F1F1} 0.811} \\ \cline{2-8} 
 & \textbf{RF} & \multicolumn{1}{c|}{\cellcolor[HTML]{EFEFEF}0.914} & \multicolumn{1}{c|}{\cellcolor[HTML]{D2D2D2}0.883} & \multicolumn{1}{c|}{\cellcolor[HTML]{C9C9C9}0.873} & \multicolumn{1}{c|}{\cellcolor[HTML]{CFCFCF}0.880} & \multicolumn{1}{c|}{\cellcolor[HTML]{BABABA}0.857} & \cellcolor[HTML]{A6A6A6}{\color[HTML]{F1F1F1} 0.836} \\ \cline{2-8} 
 & \textbf{SVM} & \multicolumn{1}{c|}{\cellcolor[HTML]{C8C8C8}0.872} & \multicolumn{1}{c|}{\cellcolor[HTML]{D1D1D1}0.882} & \multicolumn{1}{c|}{\cellcolor[HTML]{C9C9C9}0.873} & \multicolumn{1}{c|}{\cellcolor[HTML]{D6D6D6}0.887} & \multicolumn{1}{c|}{\cellcolor[HTML]{9F9F9F}{\color[HTML]{F1F1F1} 0.829}} & \cellcolor[HTML]{878787}{\color[HTML]{F1F1F1} 0.803} \\ \cline{2-8} 
\multirow{-6}{*}{\textbf{25   minutes}} & \textbf{Transformer} & \multicolumn{1}{c|}{\cellcolor[HTML]{CCCCCC}0.877} & \multicolumn{1}{c|}{\cellcolor[HTML]{BDBDBD}0.861} & \multicolumn{1}{c|}{\cellcolor[HTML]{E7E7E7}0.905} & \multicolumn{1}{c|}{\cellcolor[HTML]{E0E0E0}0.898} & \multicolumn{1}{c|}{\cellcolor[HTML]{A8A8A8}{\color[HTML]{F1F1F1} 0.838}} & \cellcolor[HTML]{A6A6A6}{\color[HTML]{F1F1F1} 0.836} \\ \hline
 & \textbf{ConvLSTM} & \multicolumn{1}{c|}{\cellcolor[HTML]{565656}{\color[HTML]{F1F1F1} 0.752}} & \multicolumn{1}{c|}{\cellcolor[HTML]{D0D0D0}0.881} & \multicolumn{1}{c|}{\cellcolor[HTML]{D6D6D6}0.887} & \multicolumn{1}{c|}{\cellcolor[HTML]{D1D1D1}0.882} & \multicolumn{1}{c|}{\cellcolor[HTML]{B3B3B3}0.850} & \cellcolor[HTML]{808080}{\color[HTML]{F1F1F1} 0.796} \\ \cline{2-8} 
 & \textbf{LR} & \multicolumn{1}{c|}{\cellcolor[HTML]{BFBFBF}0.863} & \multicolumn{1}{c|}{\cellcolor[HTML]{B1B1B1}0.848} & \multicolumn{1}{c|}{\cellcolor[HTML]{C0C0C0}0.864} & \multicolumn{1}{c|}{\cellcolor[HTML]{9B9B9B}{\color[HTML]{F1F1F1} 0.825}} & \multicolumn{1}{c|}{\cellcolor[HTML]{1D1D1D}{\color[HTML]{F1F1F1} 0.691}} & \cellcolor[HTML]{191919}{\color[HTML]{F1F1F1} 0.687} \\ \cline{2-8} 
 & \textbf{LSTM} & \multicolumn{1}{c|}{\cellcolor[HTML]{B1B1B1}0.848} & \multicolumn{1}{c|}{\cellcolor[HTML]{C1C1C1}0.865} & \multicolumn{1}{c|}{\cellcolor[HTML]{ECECEC}0.910} & \multicolumn{1}{c|}{\cellcolor[HTML]{D4D4D4}0.885} & \multicolumn{1}{c|}{\cellcolor[HTML]{B5B5B5}0.852} & \cellcolor[HTML]{8F8F8F}{\color[HTML]{F1F1F1} 0.812} \\ \cline{2-8} 
 & \textbf{RF} & \multicolumn{1}{c|}{\cellcolor[HTML]{DBDBDB}0.892} & \multicolumn{1}{c|}{\cellcolor[HTML]{CFCFCF}0.880} & \multicolumn{1}{c|}{\cellcolor[HTML]{C9C9C9}0.873} & \multicolumn{1}{c|}{\cellcolor[HTML]{C7C7C7}0.871} & \multicolumn{1}{c|}{\cellcolor[HTML]{C4C4C4}0.868} & \cellcolor[HTML]{A9A9A9}{\color[HTML]{F1F1F1} 0.839} \\ \cline{2-8} 
 & \textbf{SVM} & \multicolumn{1}{c|}{\cellcolor[HTML]{BFBFBF}0.863} & \multicolumn{1}{c|}{\cellcolor[HTML]{CDCDCD}0.878} & \multicolumn{1}{c|}{\cellcolor[HTML]{CECECE}0.879} & \multicolumn{1}{c|}{\cellcolor[HTML]{D8D8D8}0.889} & \multicolumn{1}{c|}{\cellcolor[HTML]{A7A7A7}{\color[HTML]{F1F1F1} 0.837}} & \cellcolor[HTML]{949494}{\color[HTML]{F1F1F1} 0.817} \\ \cline{2-8} 
\multirow{-6}{*}{\textbf{30   minutes}} & \textbf{Transformer} & \multicolumn{1}{c|}{\cellcolor[HTML]{C4C4C4}0.868} & \multicolumn{1}{c|}{\cellcolor[HTML]{CECECE}0.879} & \multicolumn{1}{c|}{\cellcolor[HTML]{EEEEEE}0.912} & \multicolumn{1}{c|}{\cellcolor[HTML]{E0E0E0}0.898} & \multicolumn{1}{c|}{\cellcolor[HTML]{989898}{\color[HTML]{F1F1F1} 0.821}} & \cellcolor[HTML]{979797}{\color[HTML]{F1F1F1} 0.820} \\ \hline
\end{tabular}
\end{table}

\begin{figure}
    \centering
    \includegraphics[width=\textwidth]{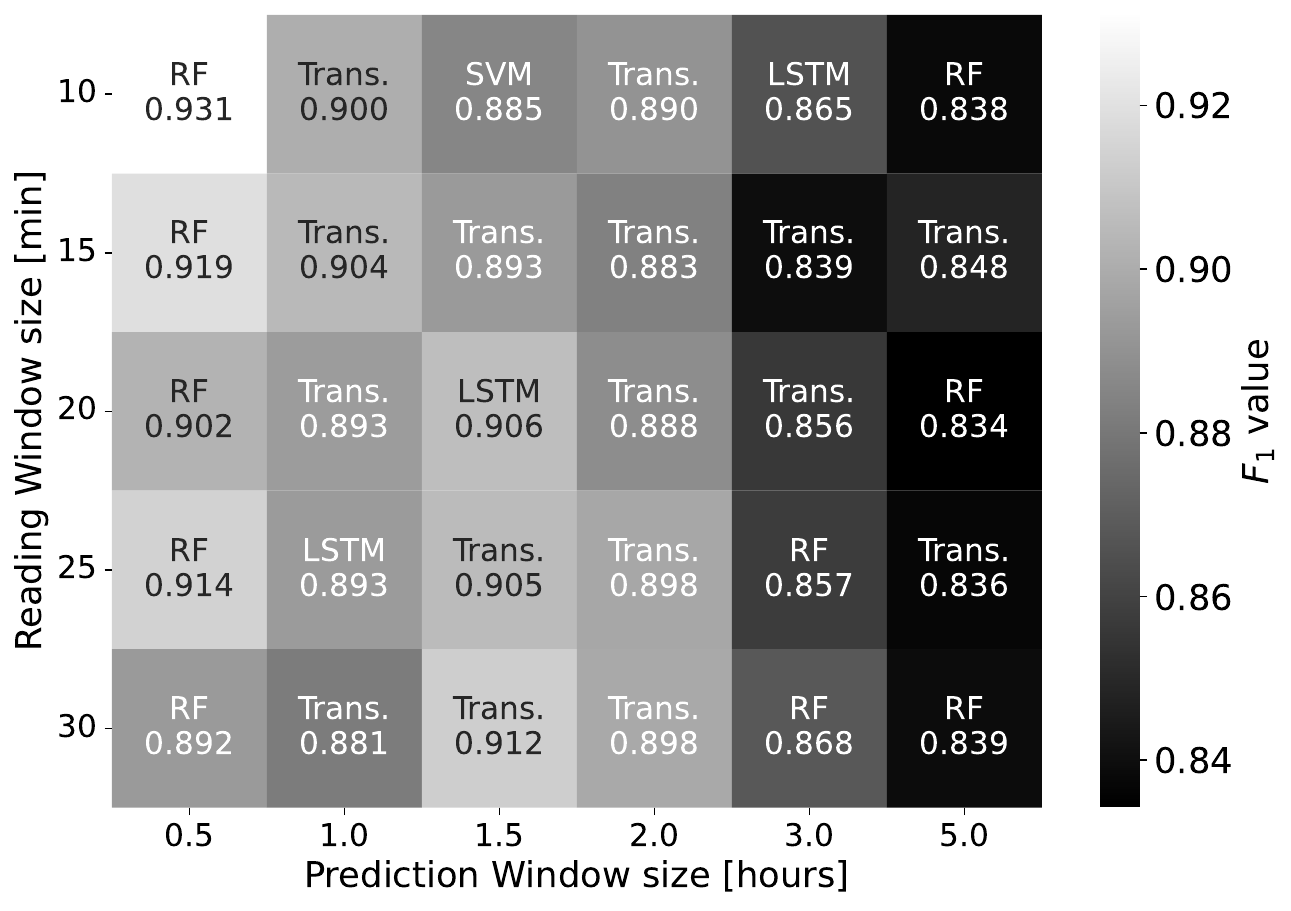}
    \caption{Comparison of the best $F_1$ scores for each combination of RW and PW sizes for the nitrogen generator case study. The heatmap shows each combination's best method and the corresponding $F_1$ score, where "Trans." indicates the Transformer. Lighter background colors correspond to better results.}
    \label{fig:nitrogen_f1_best}
\end{figure}

\textcolor{red}{
Table \ref{tab:all_f1_nitrogen} compares the $F_1$ scores of each algorithm as PW and RW vary. Also in this case, the performances of ML and DL algorithms are similar, suggesting that as the data set becomes simpler, more complex algorithms become less effective and often worse than simple ones. In particular, the best DL algorithms for each RW-PW combination reach, on average, an $F_1$ score of $89.1\%$, while the best ML algorithms for each RW-PW combination reach an average of $89.0\%$. As in the blood refrigerators case, this result can be explained considering the easily identifiable and similar anomalous patterns preceding faults, as visible in Figure \ref{fig:nitrogen_patterns}.}

\textcolor{red}{Figure \ref{fig:nitrogen_f1_best} presents the algorithm with the best $F_1$ score for each RW and PW pair. The  results highlight the similar performances of different algorithms on all the RWs and PWs. In particular, the average $F_1$ score of the best algorithm varies between $87.9\%$ and $88.5\%$ as the RW changes, a smaller difference ($\approx 0.6\%$) compared to the other data sets.}

\textcolor{red}{In this case, RF is, on average, the best algorithm (with an $F_1$ score of $\approx 87.3\%$). It benefits from simple and repetitive  patterns in the data, that it is able to capture effectively. Other algorithms have similar performances. Transformers have an average $F_1$ of $\approx 86.6\%$, LSTM has an average $F_1$ of $\approx 85.6\%$, SVM has an average $F_1$ of $\approx 85.6\%$, ConvLSTM has an average $F_1$ of $\approx 85.3\%$, and LR has an average $F_1$ of $\approx 78.4\%$.}

\section{Conclusions}
\label{sec:conclusions}
Failure prediction on industrial multivariate data is crucial for implementing effective predictive maintenance strategies to reduce downtime and increase productivity and operational time. However, achieving this goal with non-neural machine learning and deep learning models is challenging and requires a deep understanding of the input data.

\textcolor{red}{The illustrated case studies} show the importance of setting meaningful prediction and reading windows consistent with domain-specific requirements. Experimental results demonstrate that basic, general purpose  algorithms, such as  Logistic Regression already achieve acceptable performances in complex cases. \textcolor{red}{In the wrapping machine case study, Logistic Regression attains a macro $F_1$ score of 0.789 with a prediction window of 15 minutes and a reading window of 20 minutes. In the same case study, LR is outperformed by LSTM, a complex model that exploits temporal dependencies, with a  margin of more than 7\% on average on all RW-PW pairs}. In the best-case scenario of a prediction window of 15 minutes and a reading window of 20 minutes, the LSTM model achieves a notable macro $F_1$ score of 0.861. However, the performance gap declines as the prediction horizon enlarges because the relevance of the temporal dependencies between the RW and the PW  fades out. Ultimately, an LSTM model is maximally effective for short PWs, when the influence of the reading window historical data is expected to be more critical and becomes less effective as the prediction window increases.

\textcolor{red}{The better performances obtained with DL algorithms, however, are negligible in the case of simpler data sets, where easily identifiable repetitive patterns can be found. The blood refrigerator data set is a simpler data set, with a few relevant anomalous patterns leading to failures. In that case, ML algorithms perform similar to DL algorithms, often surpassing the latter. A similar result is obtained for the nitrogen generator data set. In this case, Random Forest, which has the worst performances on the wrapping machine data set, can effectively find simple rule-based repetitive patterns anticipating failures, achieving the best performances.}

The results presented in this paper are valid for the industrial scenario and data sets employed in the experiments. \textcolor{red}{Larger data sets} would likely contain more occurrences of multiple alert codes enabling the study of different types of anomalies.

Our future work will focus on  fine-tuning the LSTM model and exploring different architectures (e.g., CNN-based models \cite{IrinSherly2023} and Gaussian Processes \cite{Lee2023}) and hybrid models that combine multiple machine learning and deep learning techniques \cite{Wahid2022}. Such a comparison can help identify the most effective approach for different industrial scenarios and enable a better  understanding of how different network structures and learning algorithms impact failure prediction accuracy. Another fundamental research direction for the practical application of  LSTM  as predictors is the interpretability  of their output \cite{pmlr-v97-guo19b}. As DL models are often used as black boxes, understanding the reasons behind their predictions is crucial in   industrial settings where the model's output is used to take preventive maintenance action.

\section*{Acknowledgments}
This work has been supported by the European Union’s Horizon 2020 project PRECEPT, under grant agreement No. 958284.

\section*{Declaration of competing interest}
The authors declare that they have no known competing financial interests or personal relationships that could have appeared to influence the work reported in this paper.

 \bibliographystyle{elsarticle-num} 
 \bibliography{cas-refs}

\end{document}